%% file: sgpn_arxiv.tex
\documentclass[10pt,twocolumn,letterpaper]{article}

\usepackage{iccv}
\usepackage{times}
\usepackage{epsfig}
\usepackage{graphicx}
\usepackage{amsmath}
\usepackage{amssymb}
\usepackage{amsthm}
\usepackage{animate}
\usepackage{algorithm}
\usepackage[noend]{algpseudocode}

\usepackage[breaklinks=true,bookmarks=false]{hyperref}

\iccvfinalcopy 

\def\iccvPaperID{1882} 

\ificcvfinal\fi

\begin{document}

\title{Learning Propagation for Arbitrarily-structured Data}

\author{Sifei Liu$^{1}$,\, Xueting Li$^{1,2}$,\, Varun Jampani$^{1}$\thanks{The current affiliation is Google Research.},\, Shalini De Mello$^{1}$,\, Jan Kautz$^{1}$ \\
$^{1}$NVIDIA,\, $^{2}$University of California, Merced \\
}

\maketitle
\ificcvfinal\thispagestyle{empty}\fi

\input{abs.tex}
\input{intro_v2.tex}
\input{related.tex}
\input{approach_v2.tex}
\input{implementation_v2.tex}
\input{experiment.tex}

\input{conclusion.tex}
\clearpage
\input{supp_arxiv.tex}
{\small
\bibliographystyle{ieee_fullname}
\bibliography{sgpn}
}

\end{document}

%% file: abs.tex
\begin{abstract}
%
Processing an input signal that contains arbitrary structures, e.g., superpixels and point clouds, remains a big challenge in computer vision.
Linear diffusion, an effective model for image processing, has been recently integrated with deep learning algorithms.
In this paper, we propose to learn pairwise relations among data points in a global fashion to improve semantic segmentation with arbitrarily-structured data, through spatial generalized propagation networks (SGPN).
The network propagates information on a group of graphs, which represent the arbitrarily-structured data, through a learned, linear diffusion process. 
The module is flexible to be embedded and jointly trained with many types of networks, e.g., CNNs.
We experiment with semantic segmentation networks, where we use our propagation module to jointly train on different data -- images, superpixels and point clouds.
%
We show that SGPN consistently improves the performance of both pixel and point cloud segmentation, compared to networks that do not contain this module.
Our method suggests an effective way to model the global pairwise relations for arbitrarily-structured data.
\end{abstract}

%% file: intro_v2.tex
\vspace{-3mm}
\section{Introduction} \label{sec:intro}

The individual visual elements of spatially distributed data, \eg, pixels/superpixels in an image or points in a point cloud, exhibit strong pairwise relations. Capturing these relations is important for understanding and processing such data. 
For example, in semantic segmentation, where each pixel/point
is assigned a semantic label, it is very likely that the points
that are spatially and photometrically close, or structurally connected to each other 
have the same semantic label, compared to those that are farther away.
We can make use of such similarity cues to infer the relationships among points and improve the propagation of information (\eg, semantic labels, color \etc) between them.
This pairwise relationship modeling is often called ``affinity'' modeling. 
Evidence from psychological~\cite{Bar2004VisualOI, Sastyin2015} and empirical studies in computer vision~\cite{mottaghi2014role,divvala2009empirical,densecrf} suggests that general classification or regression problems can immensely benefit from the explicit modeling of pairwise affinities.

\begin{figure}
    \centering
    \includegraphics[width=0.99\linewidth]{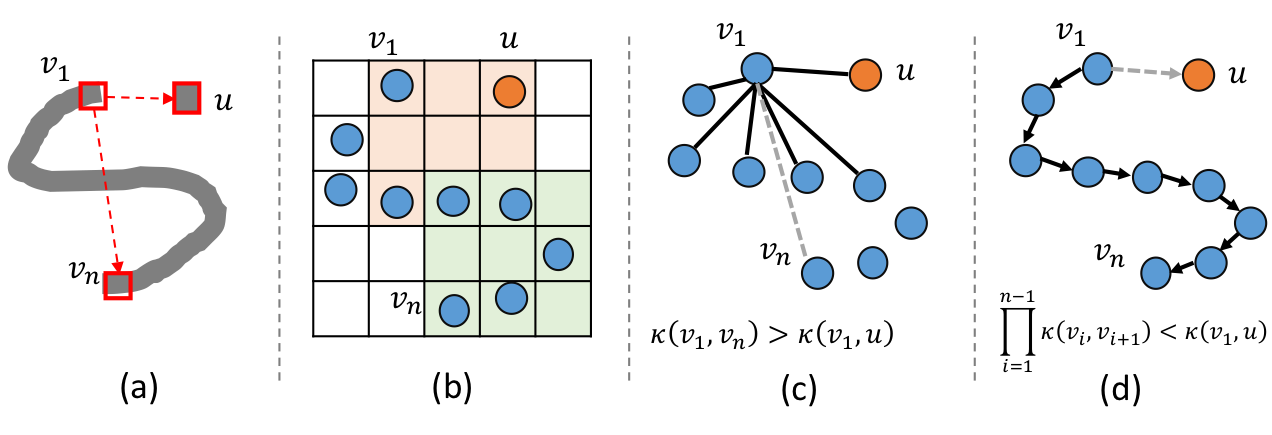}
    \caption{
    Groupings of different objects $v$ and $u$ in (a) with different strategies:
    (b) performing convolution on grids;
    explicit pairwise modeling via
    (c) fully-connected graphs, and
    (d) our path-aware propagation.
    Since $v$ and $u$ have the same color, we model the similarity ($\kappa$) using spatial closeness between two points.}
    \label{fig:nlmgpn}\vspace{-4mm}
\end{figure}

With a dramatic rise in adoption for computer vision tasks, CNNs implicitly model pairwise relationships, as convolution filters learn to capture
correlations across image pixels.
Several extensions of CNNs to process arbitrarily-structured data (such as point clouds) have been proposed (\eg, permutohedral lattice~\cite{adams2010fast,jampani2016learning, su2018splatnet}) that go beyond processing regular grid-like structured images. They transform the data to some regular structures, such that convolutional filters can be learned for them.
However, convolutions can only capture short-range pairwise relations and the filters are also content-agnostic as their weights are fixed once they are trained. As a result, we usually resort to using very deep network architectures to model all possible pairwise relations, and long-range pixel dependencies.
As an alternative, several recent works~\cite{zheng2015conditional,chen2015learning,chen2016semantic,jampani2016learning,lin2016efficient,chandra2016fast,gadde2016superpixel,liu2017learning,Wang_nonlocalCVPR2018,Maire2016AffinityCL} propose neural network modules that can explicitly model pairwise relations, resulting in considerable improvements in CNN performance for a variety of computer vision tasks.
However, most of them are designed on regularly-structured data, such as images and videos.

Despite the existence of these methods,
several important challenges remain for processing arbitrarily-structured data such as point clouds:
First, we hope such data can be represented with a more flexible structure, instead of regular-grids (such as voxel grids or permutohedral lattice), such that the original structure of the input data can be faithfully preserved.
Second, as mentioned above, we hope to explicitly model the pairwise relations among their data elements.
Third, we hope to model the pairwise relations globally, but still adhere to the structures of the input data.
Fig.~\ref{fig:nlmgpn} illustrates the above challenges, where the aim is to decide for the point $v_1$, which belongs to the curved object, whether $v_n$ and $u$ belong to the same object as $v_1$.
As shown in Fig.~\ref{fig:nlmgpn}(b), placing a curve on a grid and conducting convolution on top of it does not effectively correlate the elements.
On the other hand, with explicit pairwise modeling as shown in Fig.~\ref{fig:nlmgpn}(c), if we relate $v_1$ with the other points globally by independently computing their Euclidean distances, we will incorrectly model $v_1$ and $v_n$ as ``not similar", but $v_1$ and $u$ as ``similar'', since they are spatially closer.
Fig.~\ref{fig:nlmgpn}(c) also belongs to the non-local propagation methods
\cite{densecrf,Wang_nonlocalCVPR2018,zheng2015conditional,chandra2016fast,jampani2016learning}, which explicitly model pairwise relations via a fully-connected graph.

In this work, we aim to address all the above mentioned challenges by proposing a spatial generalized propagation network (SGPN), as illustrated in Fig.~\ref{fig:nlmgpn}(d).
Instead of transforming input points into a 
regular grid structure, we retain the original spatial structure of the data, but establish several directed acyclic graphs (DAGs) to connect adjacent points, where Fig.~\ref{fig:nlmgpn}(d) shows a top-to-bottom DAG that faithfully adheres to the curved object $v$'s structure.
With our propagation operator, the distance between $v_1$ and $v_n$ is determined by the accumulated connections of the adjacent elements between them. When the multiplication of the intermediate distances is small, we can correctly model $v_1$ and $v_n$ as belonging to the same object, even though they are spatially far away.

We show that, theoretically, our propagation mechanism is equivalent to linear diffusion.
More importantly, we propose a differentiable kernel operator such that even for DAGs, the strength of an edge between two connected nodes is learnable.
Moreover, our entire framework is a flexible deep learning building block, where the SGPN can be embedded in, and jointly optimized with any type of network, \eg, any baseline CNN for semantic segmentation.
For the same reason our propagation module, which operates on arbitrarily-structured data, \eg, point clouds, can also be easily combined with 2D CNNs that process images associated with the points, \eg, the multi-view images corresponding to point clouds.
We demonstrate the effectiveness of SGPN by applying it to different types of data, including image pixels, superpixels and point clouds, for the task of semantic segmentation.
Experimental results show that our SGPN outperforms state-of-the-art methods on semantic segmentation with all types of data and consistently improves all the baseline models by reliable margins.

%% file: related.tex
\vspace{-1mm}
\section{Related Work}
\vspace{-1mm}
\paragraph{Modeling irregularly-structured data.}
Irregular data domains refer to those that do not contain regularly ordered elements, \eg, superpixels or point clouds.
Deep learning methods that support processing irregular domains are far less than those that exist for regular domains, \eg, images and videos.
For modeling superpixels, the work of~\cite{he2015supercnn} uses superpixels inside CNNs by re-arranging them by their features.
The work of~\cite{jampani2016learning} uses a superpixel
convolution module inside a neural network, which results
in some performance improvement~\cite{Tu-CVPR-2018,jampani2018superpixel}.
In comparison, quite a few networks have been designed for point clouds \cite{pointcnn, pointnet, pointnetplusplus, Tangent, su2018splatnet}, where most target adapting CNN modules to unstructured data, instead of explicitly modeling the pairwise relationships between the points.
On the other hand, while some propagation modules~\cite{densecrf,ShapePFCN,xie2018attentional,jampani2016learning} address affinity modeling for
irregularly-structured data, they cannot address the challenge of preserving internal structures due to the non-local nature of their propagation.

\vspace{-3mm}
\paragraph{Modeling pairwise affinity.}
Pairwise relations are modeled in a broad range of low- to high-level vision problems.
Image filtering techniques including edge-preserving smoothing and image denoising \cite{aurich1995non,tomasi1998bilateral,nonlocal05,he2010guided} are some of the most intuitive examples of applying pairwise modeling to real-world applications.
The task of structured prediction~\cite{densecrf,lafferty2001conditional,felzenszwalb2010object}, on the other hand, seeks to explicitly model relations in more general problems.
Recently, many methods for modeling affinity have been proposed as deep learning building blocks~\cite{liu2017learning,Wang_videogcnECCV2018,wu2017fast,Wang_nonlocalCVPR2018,zheng2015conditional,lin2016efficient,chandra2016fast,jampani2016learning,xie2018attentional}, and several of them also propose to ``learn" affinities~\cite{liu2017learning,Wang_nonlocalCVPR2018,jampani2016learning,xie2018attentional}.
Besides these methods, diffusion theory~\cite{perona1990scale} provides a fundamental framework that relates the task of explicit modeling of pairwise relations to physical processes in the real world, where many popular affinity building blocks ~\cite{Wang_nonlocalCVPR2018,xie2018attentional,liu2017learning} can be described by it.

\vspace{-3mm}
\paragraph{Propagation networks.}
Our work is related to the recent spatial propagation networks (SPNs)~\cite{liu2017learning, cheng2018depth} for images, which learn affinities between pixels to refine pixel-level classification~\cite{liu2017learning} or regression~\cite{Xu_2018_ECCV,liu2018switchable,cheng2018depth} tasks.
SPNs model affinity via a differentiable propagation layer, 
where the propagation itself is guided by learnable, spatially-varying weights that are conditioned on the input image pixels.
SPNs have the advantage of faithfully preserving complex image structures in image segmentation~\cite{liu2017learning}, depth estimation~\cite{Xu_2018_ECCV} and temporal propagation~\cite{liu2018switchable}.
We show in the following section, that our work generalizes SPNs to arbitrary graphs, such that SPN can be viewed as a special case of our work on regular grids.
%
%
Our work is also related to recurrent neural networks (RNN) on graphs~\cite{jain2016structural, de2018simple,Liang2016SemanticOP}. 
However, unlike our work RNNs are not designed for \emph{linear} diffusion on graphs, but instead target more general problems represented as graphs.

%% file: approach_v2.tex
\begin{figure*}[t]
    \centering
    \includegraphics[width=0.90\linewidth]{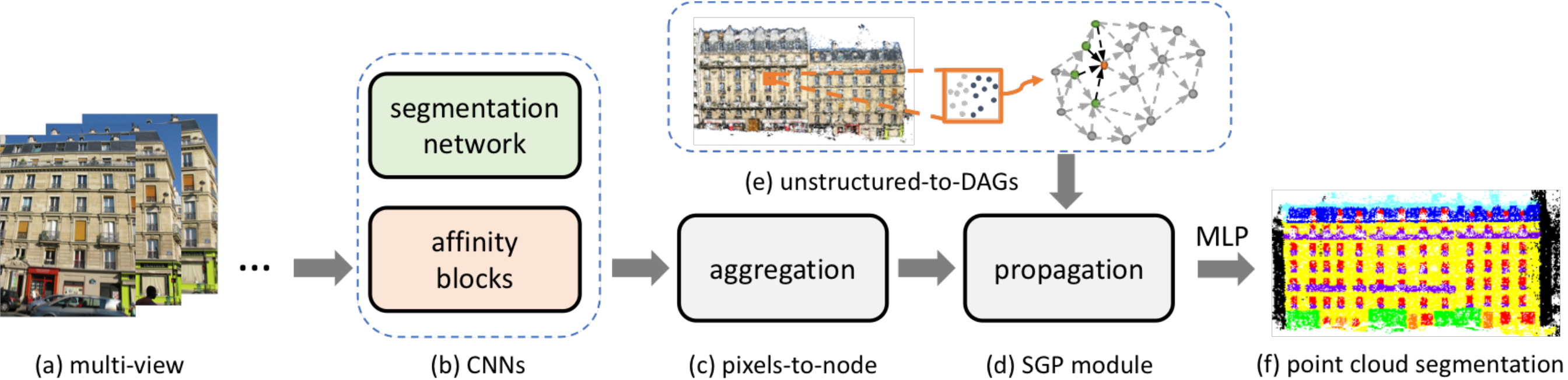}
    \caption{\footnotesize 
    A general architecture of SGPN for point cloud segmentation. See Section \ref{sec:approach} for details of individual modules. 
    }
    \label{fig:graphprop}\vspace{-3mm}
\end{figure*}

\vspace{-1mm}
\section{Spatial Generalized Propagation Network}\vspace{-1mm}
\label{sec:approach}
Unlike images where pixels are placed on regular 2D grids, data such as superpixels or point clouds encountered in vision tasks have an undefined structure.
In order to process such data with deep neural networks, they must be converted into some structures such as a high-dimensional lattice \cite{su2018splatnet} or fully connected graphs \cite{pointnet,Wang_nonlocalCVPR2018}, on which the operations of convolution, pooling, etc, can be carried out.
We take point cloud segmentation as an example in Fig.~\ref{fig:graphprop} to 
explain our method. We build a group of DAGs on the raw points, as shown in Fig.~\ref{fig:graphprop}(e), by connecting the spatially adjacent points. 
In contrast with transforming the unstructured points to a rigid lattice, where the 
topologies of the raw points may be changed, and many unused vertices may exist (e.g., see Fig.~\ref{fig:nlmgpn}(b) where many grid cells are unoccupied), DAGs neither change the input topologies, nor consume any extra memory in order to enforce a regular structure.
Such a DAG structure is highly flexible since the input points of different objects can have different DAGs exactly adhering to their own shapes.

In terms of explicit pairwise modeling, in contrast to fully connected graphs \cite{Wang_nonlocalCVPR2018} where points are densely connected (see Fig.~\ref{fig:nlmgpn}(c)), the structure of DAGs also enables the propagation along different directions to be carried out, and ``paths'' along complex shapes of the input data (\eg Fig.~\ref{fig:nlmgpn}(d)) are modeled.
We establish different directions with DAGs, \eg, along the $x$, $y$ and $z$ axis for a point cloud in 3D (6 directions in total with positive and negative directions along each axis), where we show the left-to-right DAG in Fig.~\ref{fig:graphprop}(e).
The DAGs can be built on a global scope, \eg, for a point cloud with millions of points, to support long-range propagation.


Once the DAG is constructed, we learn pairwise affinities between the DAG vertices and we use our SGPN propagation module for structured information propagation along the edges. SGPN can be attached on top of any CNN that provides initial (unary) features at DAG vertices.
In the case of point clouds, the CNN can be an existing 3D network. To demonstrate the flexibility of SGPN and to leverage the potential of 2D CNNs, we obtain vertex features using a 2D CNN on the corresponding multi-view 2D images. 
We use a differentiable aggregation module \ref{fig:graphprop}(c) that transforms the pixel features into vertex features on the DAG.
In the following part, we first describe the formulation of linear propagation on a DAG, assuming that the DAGs are given.
Then, we show that it exactly performs linear diffusion on the DAGs.
We emphasize the role of our SGPN -- to implicitly learn to relate the vertices globally and to refine the embedded representations, by learning the representation for vertices (unary) and edges (pairwise), in (Section~\ref{sec:represent}).

\vspace{-1mm}
\subsection{Formulation}\label{sec:form}
\vspace{-1mm}
\paragraph{Propagation on DAGs.}
Given a set of vertices $V=\{v_1, ..., v_N\}$ of a DAG, we denote the set of indices of the connected neighbors of $v_i$ as $K_i$. 
For example, if a DAG is built along a direction from left to right, and $V$ is a set of points in a point cloud, the vertices in $K_i$ would be the points that are adjacent to $v_i$ and are located spatially to the left of it (see Fig.~\ref{fig:graphprop}(e)).
We denote the feature of each vertex, before and after propagation, as $u\in \mathbb{R}^{N\times c}$ and $h\in \mathbb{R}^{N\times c}$, respectively, where
$u$ can be a $c$-channel feature map obtained from an intermediate layer of a segmentation CNN before propagation, and $h$ would be its value after propagation. We call $u$ and $h$ as the unary and propagated features, respectively.
The propagation operator updates the value of $h$ for the various vertices of the graph recurrently (\eg, from left-to-right) as:
\begingroup\small
\begin{equation}
    h(i) = (1-\sum_{k\in{K_i}}g_{ik})u(i) + \sum_{k\in{K_i}}g_{ik}h(k),
    \label{eq:linear}
\end{equation}\endgroup
where $\left\{g_{ik}\right\}$ is a set of learnable affinity values between $v_i$ and $v_k$, which we denote as the edge representations.

\begin{figure}[t]
    \centering
    \includegraphics[width=0.95\linewidth]{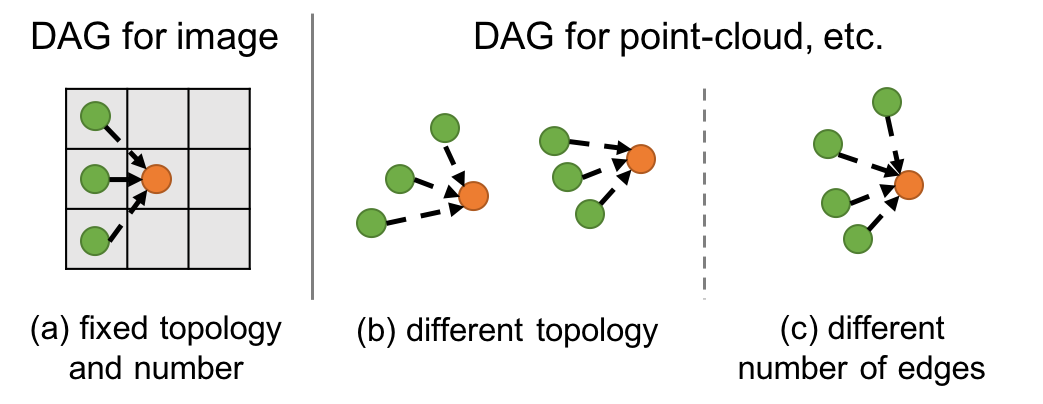}
    \caption{\footnotesize Comparisons of local connections of DAGs between (a) image pixels and (b) (c) irregularity-structured points.
    }
    \label{fig:topo}\vspace{-3mm}
\end{figure}

\vspace{-4mm}
\paragraph{A parallel formulation.} 
%
In DAGs, since vertices are updated sequentially, propagating features from one vertex to another using linear diffusion in Eq.~\eqref{eq:linear} results in poor parallel efficiency. Here we show that the propagation on DAGs can be re-formulated in a ``time-step'' manner, which can be implemented in a highly parallel manner.
This is achieved via a slight modification of the topological sorting algorithm (see Alg.~1 in the supplementary material) used to construct the DAGs: we re-order the vertices into groups to ensure that (a) vertices in the same group are not linked to each other and can be updated simultaneously, and (b) each group has incoming edges only from its preceding groups.
Taking an image as an example, we can construct a left-to-right DAG by connecting all pixels in the $t^{th}$ column to those in the $(t+1)^{th}$ column (see Fig.~\ref{fig:topo}(a)). That is, \textit{column} in an image is equivalent to a \textit{group} in a DAG, where in Eq.~\eqref{eq:linear}, pixels from the same column can be computed simultaneously.
We denote the corresponding ``unary" and ``propagated" features for the vertices in the $p^{th}$ group, before and after propagation as $u_p$ and $h_p$, respectively. We perform propagation for each group as a linear combination of all its previous groups:
\begingroup\small
\vspace{-2mm}
\begin{equation}\vspace{-2mm}
    h_p = (I-d_p)u_p+\sum_{q=1}^{p-1}w_{pq}h_q,
    \label{eq:group}
\end{equation}\endgroup
where $q$ is a group that precedes the group $p$ along the direction of propagation. Suppose the $p^{th}$ and $q^{th}$ groups contain $m_p$ and $m_q$ vertices, respectively, $w_{pq}$ is a $m_p\times m_q$ matrix that contains all the corresponding weights $\left\{g\right\}$ between vertices in $h_p$ and $h_q$.
Specifically, $d_p\in \mathbb{R}^{m_p\times m_p}$ is a diagonal degree matrix with a non-zero entry at $i$ that aggregates the information from all the $\left\{w_{pq}\right\}$ as:
\begingroup\small
\begin{equation}\vspace{-2mm}
    d_p(i,i) = \sum_{q=1}^{p-1}\sum_{j=1}^{m_q}w_{pq}(i,j).
    \label{eq:adgree}
\end{equation}\endgroup

Re-ordering vertices into groups results in the ``time-step'' form of Eq.~\eqref{eq:group}, where the update for all vertices in the same group is computed simultaneously. For one direction with the number of groups as $T$, the computational complexity for propagating on the DAG is $O(T)$. 
Given Eq.~\eqref{eq:group}, we need to explicitly maintain stability of propagation, which is described in the supplementary material.

\vspace{-3mm}
\paragraph{Diffusion on Graphs}

Linear diffusion theory states that the filtering of signals can equivalently be viewed as the solution of a heat conduction or diffusion, where the change of the signal over time can be described as spatial differentiation of the signal at the current state~\cite{perona1990scale}.
The theory can be generalized to many other processing, such as refinement of segmentation, where the spatial differentiation needs to be replaced with a task-specific Laplacian matrix.

When fitting diffusion theory into deep neural network, we hope the Laplacian to be learnable and flexibly conditioned on the input signal, through a differentiable linear diffusion module -- we achieve this goal on DAGs.
We first introduce the notations, where $U = [u_1,...,u_T]\in \mathbb{R}^{N\times c}$ and $H = [h_1,...,h_T]\in \mathbb{R}^{N\times c}$ are the features of all the $N$ ordered groups ($U$ and $H$ are re-ordered $u$ and $h$ in Eq.~\eqref{eq:linear}) concatenated together.
We re-write Eq.~\eqref{eq:group} as refining the features $U$ through a global linear transformation $H-U = -LU$.
We can derive from both Eq.~\eqref{eq:group} and Eq.~\eqref{eq:linear} that $L$ meets the requirement of being a Laplacian matrix, whose each row sums to zero. It leads to a standard diffusion process on graphs. Details of proof can be found in the supplementary material.

We note that being linear diffusion process on DAGs is an important property showing that the proposed algorithm is closely related to real physical processes widely used in many image processing techniques~\cite{perona1990scale,grady2006random,Bampis2016GraphDrivenDA}. This connection also makes our model more interpretable, for example, the edge representations $\{g_{ik}\}$ in Eq.~\eqref{eq:linear} then explicitly describe the strengths of diffusion in a local region.

\begin{figure}[t]
    \centering
    \includegraphics[width=0.95\linewidth]{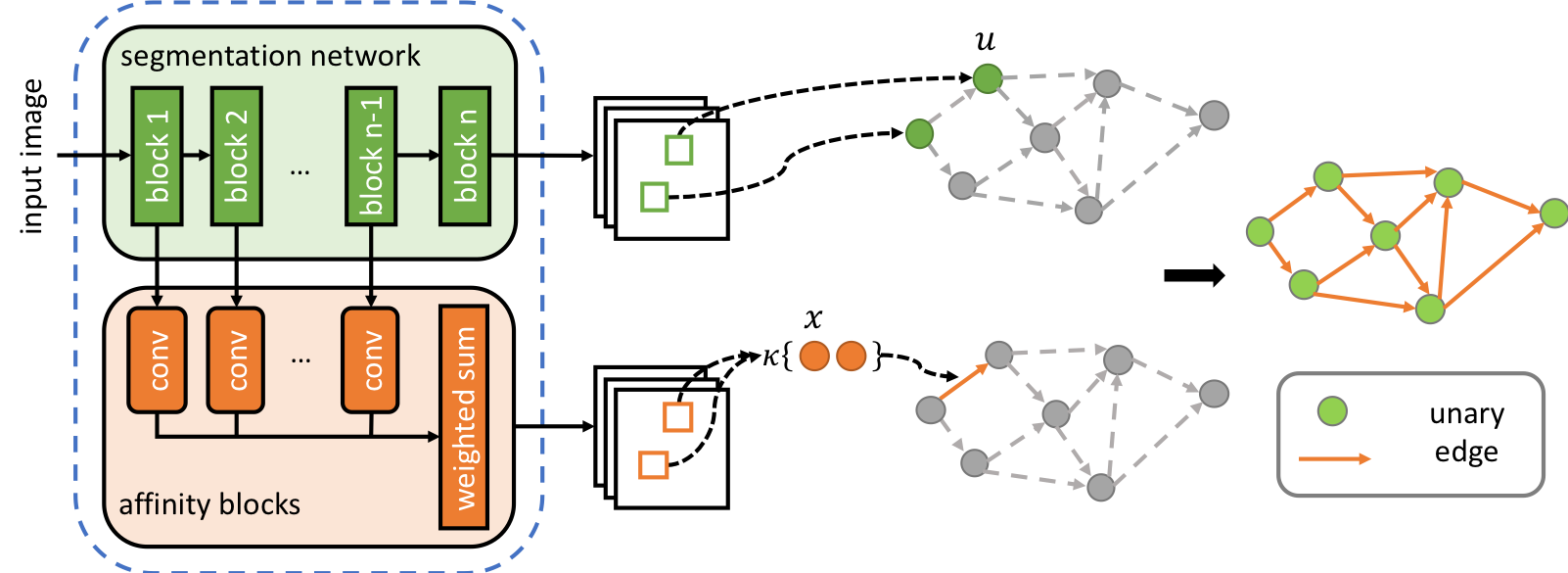}
    \caption{\footnotesize Learning the unary (green) features and the pairwise (orange) features for the edge representations of the DAG from a CNN. 
    }
    \label{fig:embed}\vspace{-3mm}
\end{figure}

\vspace{-1mm}
\subsection{Learning representations on DAGs.}\label{sec:represent}
\vspace{-2mm}
\paragraph{Learnable edge representations.}
%
An edge representation $\left\{g_{ik}\right\}$ dictates whether the value of a vertex is passed along to its neighbor or not. For the task of semantic segmentation, a desired $\{g_{ik}\}$ should represent a semantic edge (\ie, $g_{ik}=0$ stops propagation across different categories and $g_{ik}>0$ allows for propagation within a category)~\cite{chen2016semantic, CB2016Semantic,liu2016learning}.
This implies that the edge representations should be learned and conditioned on the input pixel values instead of being fixed or manually defined.
%
The work of~\cite{liu2017learning} uses the values produced by a CNN as edge representations, \ie, for left-to-right propagation, a 3-channel output is utilized to represent the edges connecting a pixel to its top-left, left, and bottom-left neighbors (Fig.~\ref{fig:topo}(a)), respectively.
However, such a method cannot generalize to arbitrarily-structured data since: (a) all vertices must have a fixed number of connected neighbors, and (b) all the connections of all pixels should have the same fixed topology or spatial layout. In contrast, here we are dealing with DAGs constructed from unstructured points (\eg, point clouds) that do not follow either of these assumptions, see Fig.~\ref{fig:topo}(b)(c).

To overcome this limitation, in our work each edge representation $g_{ik}$ used in linear propagation in Eq.~\eqref{eq:linear} is directly computed via a differentiable symmetric kernel function $\kappa$ (\eg, inner-product), such that $g_{ij} = \kappa(x_i, x_j), j\in K_i$, which is applied to the feature vectors $x_i$ and $x_j$ that are specifically computed to relate vertices $v_i$ and $v_j$.
We denote $x\in \mathbb{R}^{N\times c}$ as feature from a pairwise branch of the CNN.
Encoding the graph's edge weights in this manner, allows for each vertex to have a different 
number and spatial distribution of connected neighbors. It also reduces the task of learning edge representations $g_{ik}$ in Fig.~\ref{fig:embed} to that of learning common feature representations $\left\{x_i\right\}$ that relate the individual vertices. 
In detail, we use two types of local similarity kernels:

\vspace{-4mm}
\paragraph{Inner product (\textit{-prod}).}$\kappa$ can be defined as an inner-product similarity:
\begingroup\small\vspace{-3mm}
\begin{equation}
    \kappa(x_i, x_j) = \bar{x_i}^\top\bar{x_j}
\end{equation}\endgroup
Here $\bar{x}$ denotes a normalized feature vector, which can be computed in CNNs via Layer Normalization~\cite{Ba2016LayerN}.

\vspace{-4mm}
\paragraph{Embedded Gaussian (\textit{-embed}).} We compute the similarity in an embedding space via a Gaussian function.
\begingroup\small\vspace{-2mm}
\begin{equation}
    \kappa(x_i, x_j) = e^{-\|x_i-x_j\|_F^{2}}\vspace{-1mm}
\end{equation}
\endgroup
Since $g_{ik}$ is allowed to have negative values, we add a learnable bias term to the embedded Gaussian and initialize it with a value of $-0.5$.

\vspace{-3mm}
\paragraph{Learning Unary and Pairwise Features.}
Our network contains three blocks -- a CNN block (Fig.~\ref{fig:graphprop}(b)), that learns features from 2D images that correspond to the unstructured data (e.g., multi-view images for a point cloud, Fig.~\ref{fig:graphprop}(a)), an aggregation block (Fig.~\ref{fig:graphprop}(c)) to aggregate features from pixels to points, and a propagation (Fig.~\ref{fig:graphprop}(d)) block that propagates information across the vertices of different types of unstructured data.

We use a CNN block to learn the unary $u$ and pairwise $x$ features jointly. The CNN block can be any image segmentation network (\eg DRN~\cite{Yu2017}), where the unary term can be the feature maps before the output, or the previous upsampling layer (Fig.~\ref{fig:embed}).
Then, both features from image domain are aggregated by averaging the individual feature vectors from one local area corresponding to the same point, to the specific vertex or edge, as shown in Fig.~\ref{fig:embed}.

Since we show that the edge representations $\{g_{ik}\}$ can be computed by applying a similarity kernel to pairs of features $x_i$ and $x_j$, one could reuse the unary features (i.e., $u_i = x_i$) for computing pairwise affinities as well~\cite{Wang_nonlocalCVPR2018}. However, we find that for semantic segmentation, features from lower levels are of critical importance for computing pairwise affinities because they contain rich object edge or boundary information.
Hence, we integrate features from all levels of the CNN, with simple convolutional blocks (\eg, one CONV layer for a block) to align the feature dimensions of $\left\{x\right\}$ and $\left\{u\right\}$.
We further use a weighed-sum to integrate the feature maps from each block, where the weights are scalar, learnable parameters, and are initialized with $1$ (see the dashed box in Fig.~\ref{fig:embed}).

%% file: implementation_v2.tex
\begin{figure*}[t]
    \centering
    \includegraphics[width=0.9\linewidth]{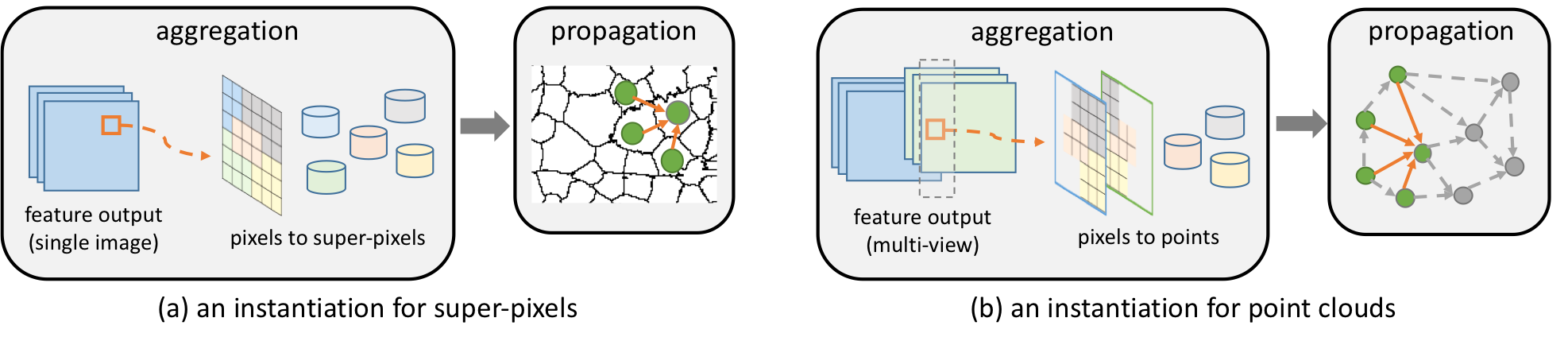}
    \caption{\footnotesize
    Different diagrams are shown for aggregation and propagation along superpixels and point clouds.
    See details in Section~\ref{sec:buildgraph}.
    }
    \label{fig:task}\vspace{-3mm}
\end{figure*}

\vspace{-2mm}
\section{Semantic Segmentation via SGPNs}\label{sec:buildgraph}\vspace{-1mm}
In this section, we introduce how to build DAGs and embed the learned representations, for refinement of semantic segmentation w.r.t different type of unstructured data.

\vspace{-1mm}
\subsection{Propagation on Pixels and Superpixels}\label{sec:supdag}
\vspace{-1mm}
\paragraph{Image.} 
We use the 3-way connection proposed in~\cite{liu2017learning} to build the DAGs for images, \ie each pixel is connected to 3 of its adjacent neighbors in each direction, and propagation is performed in all 4 directions.
Different from~\cite{liu2017learning} where the graph edge representations are directly produced by a guidance network that is separate from the segmentation network, in this work we train a single segmentation network to jointly compute both the unary features and the edge representations as the similarity between pairwise features ($x_i$).
Through this task, we demonstrate the effectiveness of our strategy for learning the edge representations, compared with~\cite{liu2017learning} as presented in Section~\ref{sec:exp}. 

\vspace{-3mm}
\paragraph{Superpixel.}
Superpixel is an effective representation to group large irregularly-shaped
semantically similar regions of an image (see Fig~\ref{fig:task}), and thus reduce the number of input elements for subsequent processing tasks. However, it is not easy to utilize superpixels directly as image pixels as they are not arranged on regular grids.
Our method can perform propagation on superpixels as an intermediate block by aggregating pixel-level features, performing propagation, and then projecting features from superpixels back to image pixels (we copy the single value of the superpixel to all the image-pixel locations that the superpixel covers).

To perform propagation, we preprocess each superpixel image by constructing a group of DAGs, where superpixels are the vertices, and the connections to their neighbors are the edges. Specifically, we search for the spatially adjacent neighbors of each superpixel, and group them into 4 groups along the 4 directions of the original image (i.e., $\rightarrow, \leftarrow,\uparrow,\downarrow$).
To determine whether a superpixel is the neighbor of another superpixel along a specific direction, we compare the locations of their centroids (see an example in Fig.~\ref{fig:task}).
For a $1024\times 2048$ image from the Cityscapes dataset~\cite{Cityscapes} with $15000$ superpixels, $T$ is around $100\sim 200$ and $200\sim 400$ for vertical and horizontal directions, respectively.
This is far more efficient than performing propagation on the original pixels of high-resolution images.

\vspace{-1mm}
\subsection{Propagation on Point Clouds}\vspace{-1mm}
Unlike many prior methods~\cite{pointnet,pointcnn} which learn features from raw points, our method flexibly maps image features to points, for which numerous off-the-shelf network architectures and pretrained weights can be utilized directly by point clouds.
%
The joint 2D-3D training is conducted by establishing the correspondences between pixels and points via camera parameters (not the focus of this work), and aggregating features from CNNs to DAGs according to the correspondences. Note that the same point may correspond to pixels from multiple images (Fig.~\ref{fig:task}(b) dashed box), where we simply average the features across them.
The construction of DAGs is similar to that of superpixels, except that the neighborhoods can be determined directly according to spatial distances between points. 

\vspace{-4mm}
\paragraph{Constructing DAGs Along Surfaces.}
We observe that constructing the graphs according to local object/scene surfaces, instead of XYZ Euclidean space, yields better performance (Section~\ref{sec:exp}). This is consistent with the intuition that local regions belonging to the same smooth and continuous surface are more likely to come from the same object.
Similar observations have been made in~\cite{SurfConv,Tangent,pointnetplusplus}. In detail, consider a set of neighboring points $k\in K_i$ in a spherical range of $i$, such that $\|\mathbf{P}(i)-\mathbf{P}(k)\|<R$. The distance between $i$ and $k$ is computed as $(\mathbf{P}(i)-\mathbf{P}(k))\cdot \mathbf{n}(i)$, where $\mathbf{P}(i)$ denotes the world coordinates of $i$, and $\mathbf{n}(i)$ is the surface normal. A subset of neighbors with the smallest distances are selected, which are equivalent to a neighborhood in the Tangent space~\cite{Tangent}. 

\vspace{-4mm}
\paragraph{Geometric-aware Edge Representations.}
Aside from learning image pixel features, we found that incorporating geometry hints for each point is equally important~\cite{su2018splatnet}. The geometry information is the concatenation of point XYZ, surface normal and point color RGB in our work.  
We map this vector from point to pixels according to the correspondence indices to form a 9-channel input map with the same resolution as the input image, and apply one single Conv+ReLU unit before integrating them with the affinity block. To avoid absolute coordinates (e.g., X are around 50 in the training set, but 90 in the validation set), we replace $(X,Y,Z)$ with the coordinate-gradient $(dX,dY,dZ)$.

%% file: experiment.tex
\begin{figure*}[th]
\centering
\includegraphics[width=0.85\linewidth]{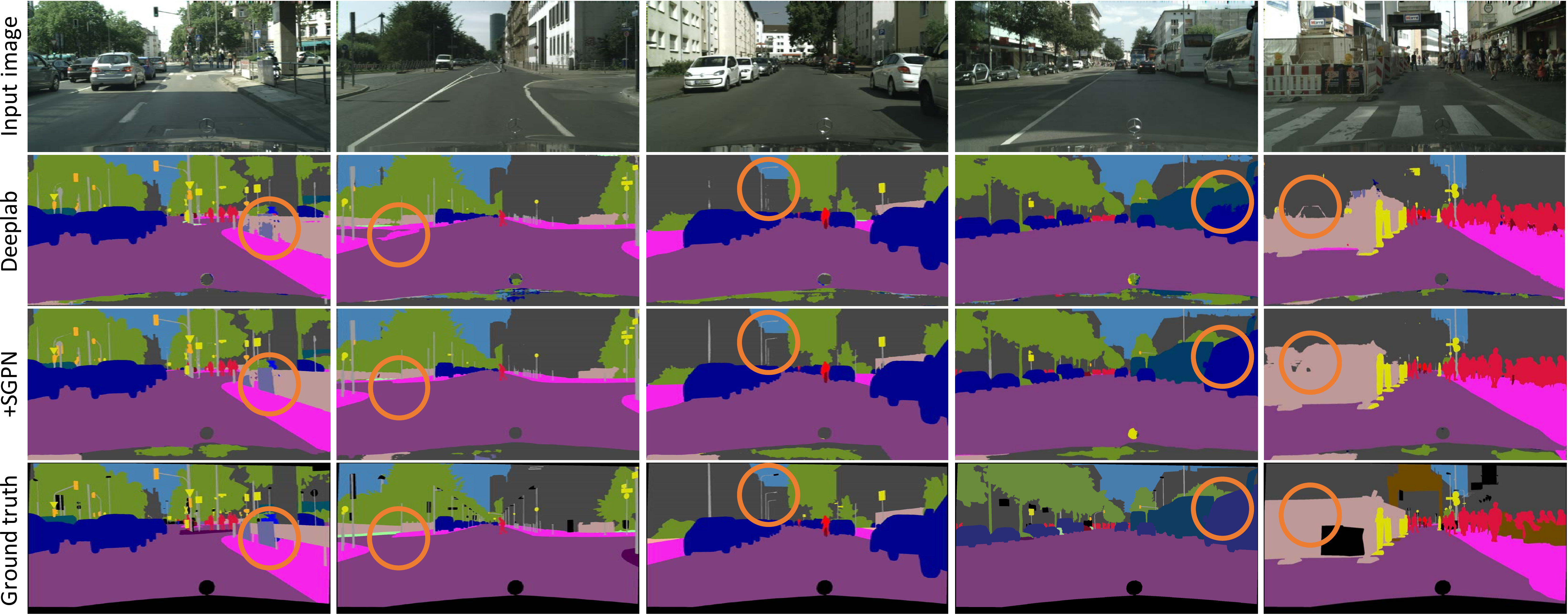}
\caption{Semantic segmentation results on the Cityscapes dataset via the \textbf{Deeplab} network. Regions are circled for ease of comparison.
}
\label{fig:result_img}
\end{figure*}

\vspace{-2mm}
\section{Experimental Results}\label{sec:exp}\vspace{-1mm}
We evaluate the performance of the proposed approach on the task of image semantic segmentation, in Sections~\ref{sec:pixel} and \ref{sec:superpixel}, and point cloud segmentation in Section~\ref{sec:point}.

\vspace{-1mm}
\subsection{Datasets and Backbone Networks}\vspace{-2mm}
%

\textbf{Cityscapes~\cite{Cityscapes}.} This dataset contains 5000 high quality, high-resolution images finely annotated with street scenes, stuff and objects, in total with 19 categories.
We use the standard training and validation sets.
For all experiments, we apply random cropping, re-scaling (between the ratio of 0.8 to 1.3), rotation ($\pm 10$ degrees) and mirroring for training.
We do not adopt any other data augmentation, coarse annotations and hard-mining strategies, in order to 
analyze the utility of the propagation module itself.

\textbf{RueMonge2014~\cite{facade}.} This dataset provides a benchmark for 2D and 3D facade segmentation, which contains 428 multi-view, high-resolution images, and a point cloud with approximately 1M 3D points that correspond to these images.
The undetected regions are masked out and ignored in processing.
Semantic segmentation labels for both image pixels and points are provided for a total of 7 categories.
Our experiments use standard training and validation splits and the same data augmentation methods as the Cityscapes dataset.

\begin{table*}[h]\setlength{\tabcolsep}{3.6pt}
    \centering
    \caption{Results for \textbf{DRN}-related networks for semantic segmentation on Cityscapes val set. We show the results of multi-scale testing, except for the +SPN, which shows better performance via single scale setting. \emph{embed} and \emph{prod} denote the embedded Gaussian and inner product kernels.} \label{tab:drn-pixel}
    {\footnotesize
    \begin{tabular}{l|lllllllllllllllllll|l}
         categories & \rotatebox[origin=c]{90}{road} & \rotatebox[origin=c]{90}{sidewalk} & \rotatebox[origin=c]{90}{building} & \rotatebox[origin=c]{90}{wall} & \rotatebox[origin=c]{90}{fence} & \rotatebox[origin=c]{90}{pole} & \rotatebox[origin=c]{90}{trafficlight} & \rotatebox[origin=c]{90}{trafficsign} & \rotatebox[origin=c]{90}{vegetation} & \rotatebox[origin=c]{90}{terrain} & \rotatebox[origin=c]{90}{sky} & \rotatebox[origin=c]{90}{person} & \rotatebox[origin=c]{90}{rider} & \rotatebox[origin=c]{90}{car} & \rotatebox[origin=c]{90}{truck} & \rotatebox[origin=c]{90}{bus} & \rotatebox[origin=c]{90}{train} & \rotatebox[origin=c]{90}{motorcycle} & \rotatebox[origin=c]{90}{bicycle} & mIoU \\\hline
        Baseline (DRN)~\cite{Yu2017} & 97.4 & 80.9 & 91.1 & 32.9 & 54.9 & 60.6 & 65.6 & 75.9 & 92.1 & 59.1 & 93.4 & 79.2 & 57.8 & 92.0 & 42.9 & 65.2 & 55.2 & 49.4 & 75.2 & 69.5 \\
        +SPN (single)~\cite{liu2017learning} & 97.7 & 82.7 & 91.3 & 34.4 & 54.6 & 61.8 & 65.9 & 76.4 & 92.2 & 62.2 & 94.4 & 78.5 & 56.8 & 92.7 & 47.8 & 70.2 & 63.6 & 52.0 & 75.3 & 71.1 \\
        +NLNN~\cite{Wang_nonlocalCVPR2018} & 97.4 & 81.1 & 91.2 & 43.0 & 52.9 & 60.3 & 66.1 & 74.9 & 91.7 & 60.6 & 93.4 & 79.2 & 57.7 & 93.3 & 54.4 & 73.5 & 54.7 & 54.2 & 74.4 & 71.3 \\\hline
        +SGPN-embed & 98.0 & 83.8 & 92.2 & 48.5 & \textbf{59.7} & 64.1 & 70.1 & 79.4 & 92.6 & \textbf{63.8} & \textbf{94.7} & 82.0 & 60.7 & 94.9 & 62.5 & 77.7 & 51.1 & \textbf{62.8} & \textbf{77.6} & 74.5 \\
        +SGPN-prod & \textbf{98.1} & \textbf{84.4} & \textbf{92.2} & 51.8 & 56.5 & \textbf{65.8} & \textbf{71.2} & \textbf{79.4} & \textbf{92.7} & 63.2 & 94.3 & \textbf{82.7} & \textbf{65.1} & \textbf{94.9} & \textbf{73.8} & 78.0 & 43.2 & 59.7 & 77.4 & \textbf{75.0} \\\hline
        +SGPN-superpixels & 97.6  & 82.4 & 91.0 & \textbf{52.7} & 52.9 & 58.4 & 66.1 & 75.9 & 91.8 & 62.2 & 93.6 & 79.4 & 58.2 & 93.3 & 62.4 & \textbf{79.7} & 57.1 & 60.2 & 75.1 & 73.2\\\hline
    \end{tabular}
    }\vspace{-3mm}
\end{table*}

%
Our experiments use two type of backbone networks. To compare against the baseline methods, mean Intersection over Union (IoU) is used as the evaluation metric.

\textbf{Dilated Residual Networks (DRN).} We use DRN-22-D, a simplified version of the DRN-C framework~\cite{Yu2017} as our primary backbone architecture.
This network contains a series of residual blocks, except in the last two levels, each of which is equipped with dilated convolutions.
%
The network is light-weight and divided into 8 consecutive levels and the last layer outputs a feature map that is $8\times$ smaller than the input image. 
One $1\times 1$ convolution and a bilinear upsampling layer is used after it to produce the final segmentation probability map.
We use the network pretrianed on  ImageNet~\cite{Deng2009ImageNetAL}.
To make the settings consistent between the different experiments, we append our SGPN module to the output of level-8 of the DRN model.

\textbf{Deeplab Network.}
We adopt the Deeplab~\cite{deeplabv3plus2018} framework by replacing the original encoder with the architecture of a wider ResNets-38~\cite{widerres} that is more suitable for the task of semantic segmentation.
The encoder is divided into 8 levels and we append the SGPN to level-8.

\vspace{-1mm}
\subsection{Image Segmentation: SGPN on Pixels}\label{sec:pixel}\vspace{-1mm}
\textbf{Propagation with DRN.}
We perform pixel-wise propagation of the output of the DRN network, and compare it to its baseline performance.
We re-train the baseline DRN model with the published default settings provided by the authors~\cite{Yu2017} and obtain the mean IoUs of $68.34$ and $69.17$, for single and multi-scale inference, respectively, on the validation set. 
For SGPN, in the pairwise block, we use features for each level except the last one, and the features from the first convolution layer.
We call the features from levels 1 to 3 as lower level features, and 5 to 7 as higher level ones. 
For the lower level features, the pairwise block contains a combination of CONV+ReLU+CONV, 
while for the higher level features, we use a single CONV layer since they have the same resolution as the output of the encoder.
The lower and higher level features are added together to form the final pairwise features, with 128 channels.
%
%
In addition, we use two deconvolution layers on the unary and pairwise feature maps to upsample them with a stride of 2, and convert them into 32 channels.
Propagation is conducted on the $2\times$ upsampled unary features with compressed (32) channels.
As mentioned before, we use the same connections between pixels as~\cite{liu2017learning} in the propagation layer.
The feature maps produced by the propagation layer are further bilinearly upsampled to the desired resolution.
To better understand the capability of the SGPN module, we adopt the same loss function (i.e., Softmax cross-entropy), optimization solver and hyper-parameters in both the baseline and our model. 

\begin{table*}[t]\setlength{\tabcolsep}{4pt}
    \centering
    \caption{Results for point cloud segmentation on the RueMonge2014~\cite{facade} val set.
    ``image to points'' is the direct mapping of 2D segmentation results to 3D points;
    ``+method'' is short for ours+method;
    ``PF'' denotes the pairwise features from 2D image;
    ``PG'' is the pairwise features with geometry-aware input;
    ``TG'' is short for tangent and ``EU'' is short for Euclidean.
    }
    \label{tab:facade}
    {\footnotesize
    \begin{tabular}{l|ll| ll | llllll}\hline
         & \multicolumn{2}{c|}{image segmentation} & \multicolumn{2}{c|}{image to points} & \multicolumn{6}{c}{point-cloud segmentation}  \\\hline
        method & SplatNet$_{2D}$ & ours DRN & SplatNet$_{2D}$ & ours DRN & SplatNet$_{2D3D}$ & +CONV-1D & +PF/TG & +PG/TG & +PF+PG/EU & +PF+PG/TG \\\hline
        mean IoU (\%) & 69.30 & 68.17 & 68.26 & 69.16 & 69.80 & 70.35 & 72.19 & 72.43 & 72.16 & \textbf{73.66} \\\hline
    \end{tabular}}\vspace{-3mm}
\end{table*}

\textbf{Comparison with SPN~\cite{liu2017learning}.}
We produce a coarse segmentation map for each training image using the baseline network mentioned above. We then re-implement the SPN model in~\cite{liu2017learning} and train it to refine these coarse segmentation maps.
The SPN shows obvious improvements over the baseline model with mIoUs of $71.1$ and $70.8$ for single and multi-scale implementations, respectively.
However, the SPN may not equally improve different models, because the edge representation is not jointly trained with the segmentation network, e.g., the multi-scale implementation does not consistently outperform the single-scale one.

\textbf{Comparison with NLNN~\cite{Wang_nonlocalCVPR2018}.}
We compare with the NLNN -- one of the most effective existing modules for learning affinity (Fig \ref{fig:nlmgpn}(c)). We implement this method to make it comparable to ours, by using the same pairwise-blocks as ours for computing affinities, but by replacing the propagation layer with the non-local layer (see details in~\cite{Wang_nonlocalCVPR2018}).
This method achieves reasonably higher performance (mIoU: 71.3) versus the baseline method and is also comparable to the SPN method.

Among all the others, our SGPN method, with different kernels (Section~\ref{sec:represent})  produces significantly better results with the final mIoU of $75$, with most categories been obviously improved, as shown in Table~\ref{tab:drn-pixel}.

\begin{table}[h]
    \centering
    \caption{Results for \textbf{Deeplab} based networks for Cityscapes image semantic segmentation on the val set.}
    \label{tab:deeplab}
    {\footnotesize
    \begin{tabular}{l|l|l|l}
         mean IoU (\%) & Baseline~\cite{deeplabv3plus2018} & +SGPN-embed & +SGPN-prod \\\hline
         single-scale & 78.20 & 80.12 & 80.09 \\
         multi-scale & 78.97 & 80.42 & \textbf{80.90} \\\hline
    \end{tabular}}\vspace{-3mm}
\end{table}

\textbf{Propagation with Deeplab Network.}
We embed the SGPN into a Deeplab-based network to show that it also improves the performance of a superior base network.
%
We demonstrate the significant improvement achieved by our models, as  measured by the mIoU as in Table~\ref{tab:deeplab}. The complete table of results can be found in the supplementary material.
Note that  SPN does not show any gain on Deeplab, and NLNN consumes a large amount of GPU memory since a fully-connected graph needs to be constructed (i.e., an $N\times N$ matrix), and thus cannot be directly applied to large networks such as Deeplab.

\vspace{-2mm}
\subsection{Image Segmentation: SGPN on Superpixels}\label{sec:superpixel}\vspace{-1mm}

\begin{figure}[t]

\centering

\begin{tabular}{c@{\hspace{0.01\linewidth}}c@{\hspace{0.01\linewidth}}c@{\hspace{0.01\linewidth}}c@{\hspace{0.01\linewidth}}}
		\includegraphics[height = .27\linewidth, width = .23\linewidth]{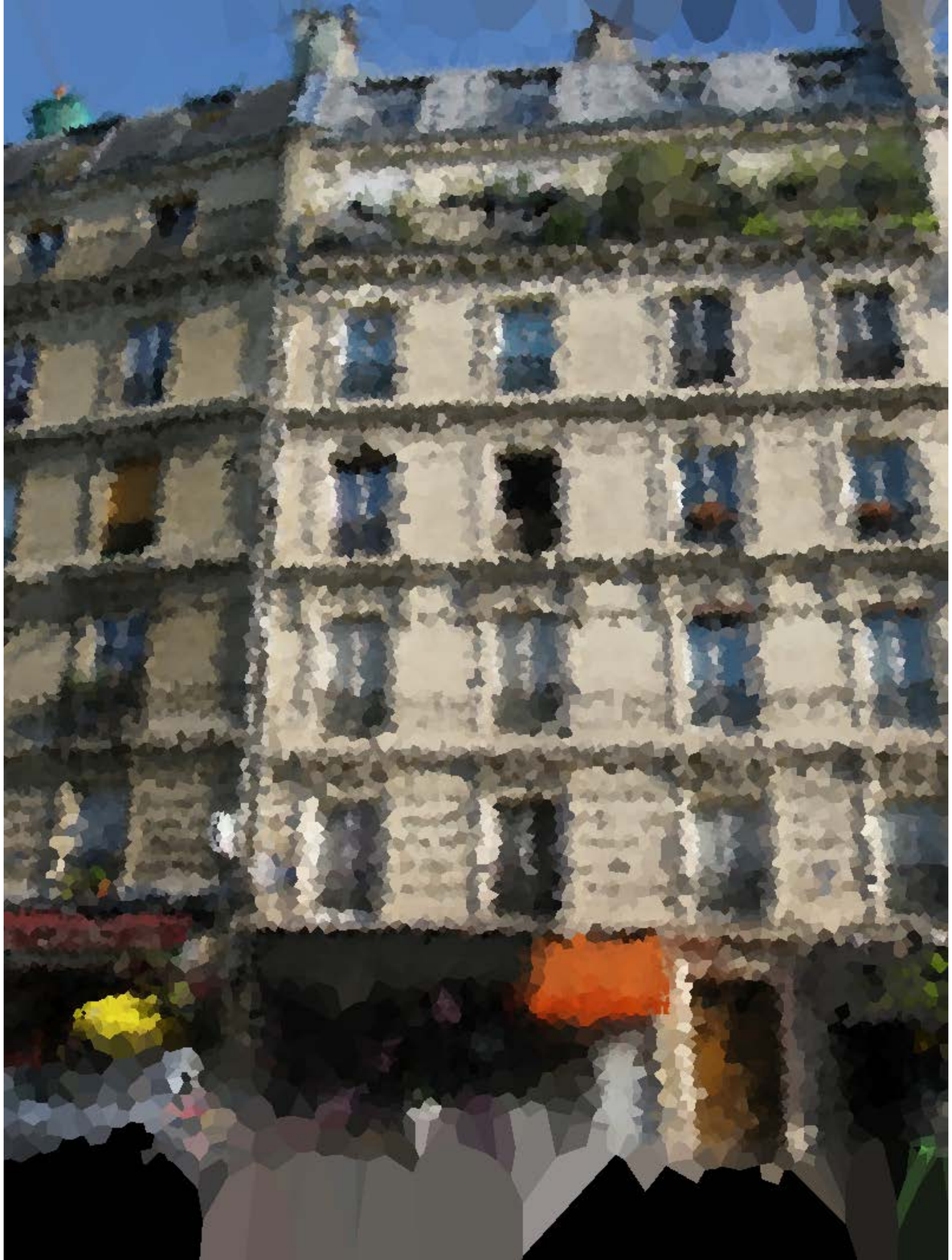} &
    	\includegraphics[height = .27\linewidth, width = .23\linewidth]{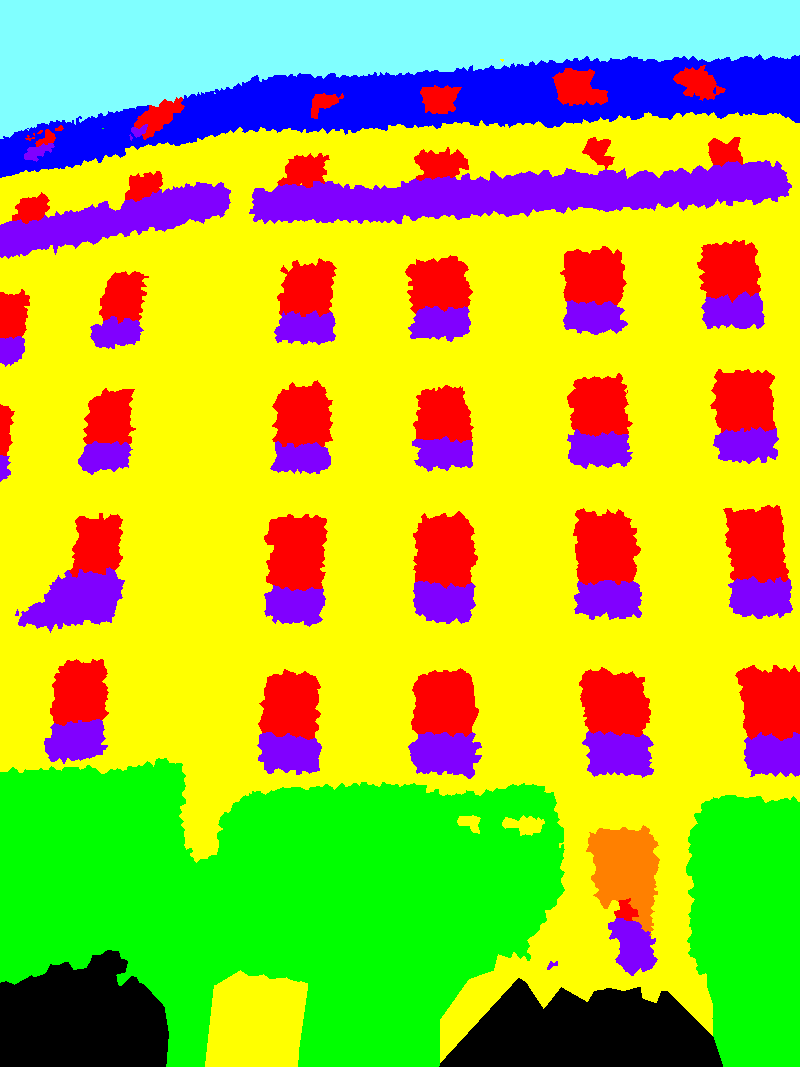} &
    	\includegraphics[height = .27\linewidth, width = .23\linewidth]{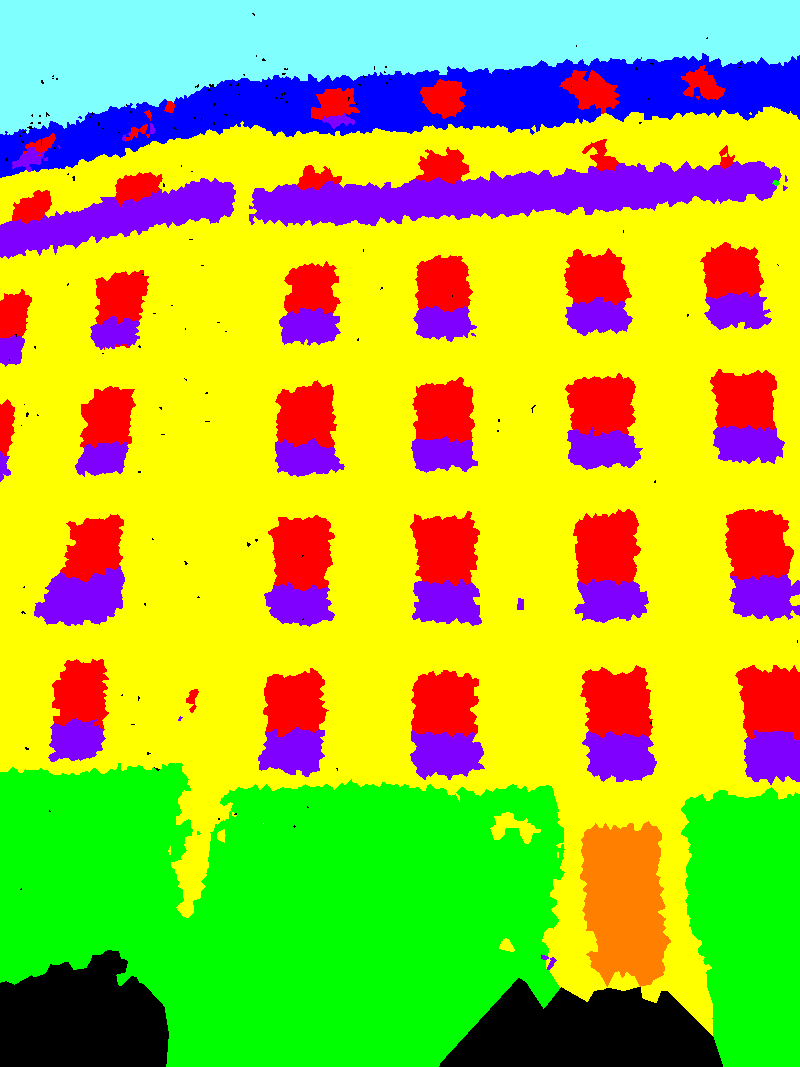} &
    	\includegraphics[height = .27\linewidth, width = .23\linewidth]{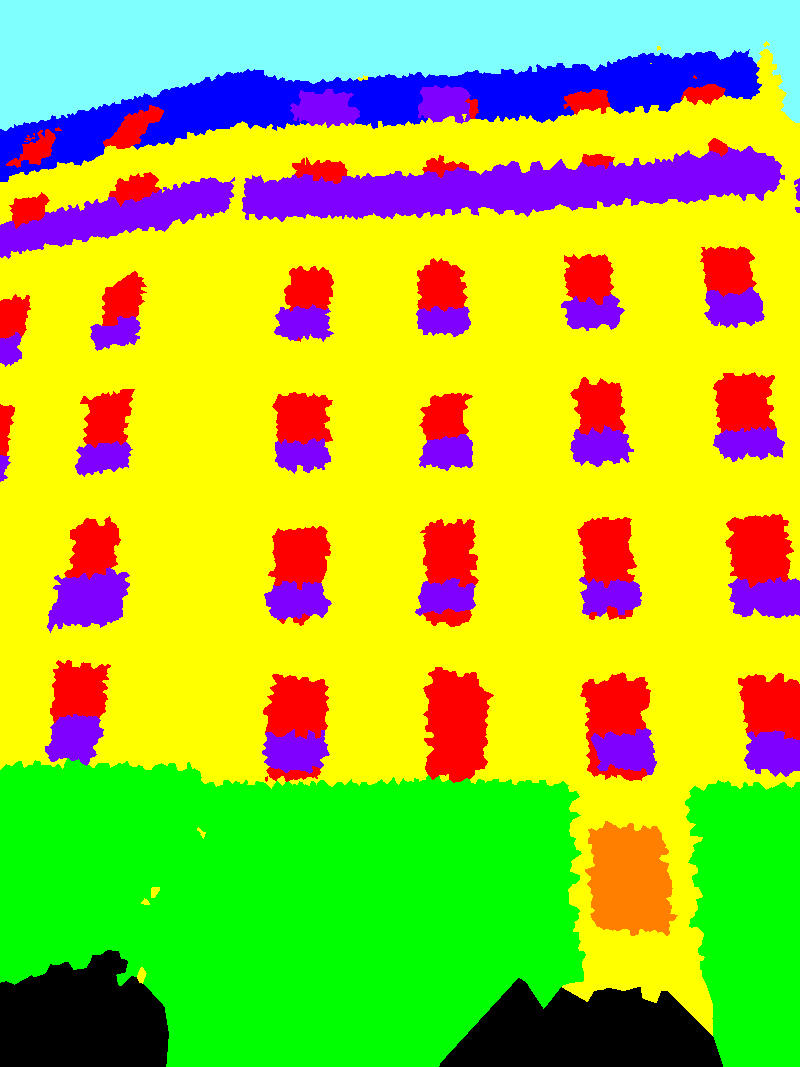} \\
    	(a) input & (b) DRN & (c) SGPN & (d) GT\\
	\end{tabular}
\caption{Qualitative comparison visualized by points to image mapping. ``DRN" is for  the direct mapping of results from the DRN image segmentation to points, ``SGPN" is for our method.}
\label{fig:whatCom}
\vspace{-3mm}
\end{figure}

We implement SGPN with superpixels (15000 per image) created by the entropy rate superpixel method \cite{Liu2011EntropyRS}. For this experiment, we use the same design as the SGPN network for pixels with the DRN backbone, but replace the pixel propagation layer with the superpixel propagation layer.
Since the model still performs pixel-wise labeling, we can directly compare it with the other DRN-related networks, as shown in Table~\ref{tab:drn-pixel}. Our SGPN with superpixel propagation shows considerable performance boost over the baseline model.
The results demonstrate that SGPN introduces an effective way of utilizing superpixels, which are generally difficult for deep learning methods to process.

\vspace{-1mm}
\subsection{Semantic Segmentation on Point Clouds}\label{sec:point}\vspace{-1mm}

\textbf{Implementation.}
Processing of point clouds for facade segmentation is different from that of superpixels: while each image corresponds to one superpixel image, in this task, many images correspond to a single point could.
Therefore, we do not need different DAGs for different images, instead, a single group of DAGs for the point is constructed.
During training, each mini-batch contains a group of sampled patches from the multi-view images, where both the unary and pairwise feature maps across samples are aggregated and mapped on to the globally constructed DAGs. 
During testing, we perform propagation on the entire validation set, by obtaining the unary and pairwise feature maps from the CNN blocks of all images, and aggregate them on the entire point cloud.
%
We use both 2D and 3D ground-truth semantic labels as the supervision signals.

\textbf{Comparison with Baselines and SOTA.}
We use the DRN-22-D as the CNN block.
To make fair comparisons against state-of-the-art work~\cite{su2018splatnet}, we evaluate the performance of the DRN for the task of multi-view image segmentation.
One direct way is to aggregate the results of image segmentation and map them on to 3D points (image to point in Table~\ref{tab:facade} ).
In addition, we jointly train the CNN block and the propagation module by adding a single $1\times 1$ 1D CONV layer before the output.
Table~\ref{tab:facade} shows the performance of the baseline models.
Our DRN model shows comparable performance to \cite{su2018splatnet} on both image labeling and point labeling with direct aggregation of features from multiple images (see column 1 and 2 in Table~\ref{tab:facade}).
The baseline model with the joint training strategy, denoted as ``+CONV-1D'', obtains the best results and outperforms the state-of-the-art method \cite{su2018splatnet} (the SplatNet$_{2D3D}$ in Table~\ref{tab:facade}), which is not jointly trained with 2D and 3D blocks.

\textbf{Ablation Study.}
%
We show the performance of our proposed approach with (a) geometric information as an additional input stream for edge representation learning, and (b) using the Tangent space to construct the DAGs (PF+PG/TG in Table~\ref{tab:facade}) shows the best results compared to the baseline and state-of-the-art methods \cite{su2018splatnet}.
To understand the contributions of individual factors, we carry out two ablation studies.
First, we compare various input streams for learning the edge representation, by removing either the geometry information (Section~\ref{sec:point}), or the image pairwise features, from the CNN block (See +PF and +PG in Table~\ref{tab:facade}).
When removing the image stream, we use an independent CNN block using the geometry input to produce the pairwise feature maps.
Second, we compare models with the same input settings for learning the edge representation, but using different ways to construct the DAGs, i.e., constructing neighborhoods via the Euclidean or Tangent spaces (Section~\ref{sec:point}) (See +PF+PG/EU and +PF+PG/TG in Table~\ref{tab:facade}).
The results demonstrate that, by solely applying the semantic feature representation learned from 2D images, or the geometry information, we can still obtain much higher mIoU compared to all of the baseline models.
However, utilizing both of them yields the best performance.
It indicates that both factors are essential for guiding the propagation on point clouds.
On the other hand, constructing the DAGs for point clouds along the Tangent space shows a significant advantage over the Euclidean space.

%% file: conclusion.tex
\vspace{-2mm}
\section{Conclusion}\vspace{-2mm}
In this work, we propose SGPN that models the global affinity for data with arbitrary structures, including but not limited to superpixels and point clouds. 
The SGPN conducts learnable linear diffusion on DAGs, with significant advantage of representing the underlying structure of the data. With the module, our method constantly outperforms state-of-the-art methods on semantic segmentation tasks with different types of input data.


%% file: supp_arxiv.tex
\section*{\Large{Appendix}}
\begin{figure*}
    \centering
    \begin{tabular}{c@{\hspace{0.005\linewidth}}c@{\hspace{0.005\linewidth}}c}
    \animategraphics[autoplay,loop,height = .15\linewidth, width = .8\linewidth]{20}{images/topogroup/}{0003}{0113} &
    \includegraphics[height = .15\linewidth]{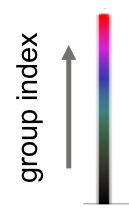}
	\end{tabular}
    \caption{{\color{red}Click to play the animation}: the dynamics of the propagation process (from the $1^{st}$ to the $n^{th}$ group) on a bottom-to-top DAG on the point cloud for buildings, in the RueMonge~\cite{facade} dataset.
    The points are ordered into 1136 groups in this case, on a global scale.
    }
    \label{fig:topogroup}
\end{figure*}

We provide additional details that are not included in the paper due to a limitation on the number of pages. These include the a mathematical proof of SGPN being equivalent to linear diffusion on graphs (Section~\ref{sec:diffuse}), the computational and experimental settings, and run-times (Section~\ref{sec:settings}), the method for re-organization of vertices (Section~\ref{sec:top}), processing of point clouds (Section~\ref{sec:point}), and additional experimental results (Section~\ref{sec:exp}). 

\section{Diffusion on Graphs: Detailed Proof}\label{sec:diffuse}
As described in the paper, with a parallel formulation, we denote the corresponding ``unary" and ``propagated" features for the vertices in the $p^{th}$ group, before and after propagation as $u_p$ and $h_p$, respectively. We perform propagation for each group as a linear combination of all its previous groups:
\begingroup\small
\vspace{-2mm}
\begin{equation}\vspace{-2mm}
    h_p = (I-d_p)u_p+\sum_{q=1}^{p-1}w_{pq}h_q.
    \label{eq:group}
\end{equation}\endgroup
Also from the paper $H = [h_1,...,h_T]\in \mathbb{R}^{N\times c}, U = [u_1,...,u_T]\in \mathbb{R}^{N\times c}$ are the features of all groups concatenated together.
By expanding the second term of Eq.~\eqref{eq:group}, the propagation can be re-written as refining the feature $U$ through a global linear transformation $H-U = -LU$. 
To meet the requirement of a linear diffusion process, $L$ should be a Laplacian matrix with the summation of the entries of each row equal to zero.
Suppose we reduce Eq.~\eqref{eq:group} to one-step (denoted as $t=1$) propagation where $h_p$ only links to $h_{p-1}$.
Then the propagation is similar to the form in \cite{liu2017learning}:
\begingroup\small
\begin{equation}
    H = (I-D_t+A_t)U, \quad t\in\{1\},
    \label{eq:affinity}
\end{equation}\endgroup
%
where $D$ is a degree matrix for the entire feature set, which places all $\{d\}$ along its diagonal and $A$ is the affinity matrix which expands the $\{w\}$ and describes the affinities among all pairs of vertices. We use $t$ to denote time steps.

By rewriting Eq.~(3) in the paper (see the definition of $m_q$ in the paper), the degree matrix between two adjacent groups can be regarded as the summation of the individual degree matrix for different time steps (we use $x(:)$ to denote all the entries in $x$): 
\begingroup\small
\begin{equation}
    d_p = \sum_{q=1}^{p-1}\sum_{j=1}^{m_q}w_{pq}(:,j) =  \sum_{q=1}^{p-1}d_{pq}.
    \label{eq:degree}
\end{equation}\endgroup
Correspondingly, by replacing $q$ with $t$ to represent the global affinity matrix of different steps, the matrix formulation is:
\begingroup\small
\begin{eqnarray}
    &D = \sum_{t}D_t, \quad t\in \{1,2,3,...\},\label{eq:affinitym}\\\nonumber
    & where\quad D_tE = A_tE.
    \vspace{-2mm}
\end{eqnarray}\endgroup
The term $A_tE$ is the expansion of $\sum_{j=1}^{m_q}w_{pq}(:,j)$.
Given Eq. \eqref{eq:affinitym}, we can re-formulate Eq.~\eqref{eq:group} as the summation of multiple time steps:
\begingroup\small
\begin{equation}
    H = (I-\sum_{t}D_t+\sum_{t}A_t)U, \quad t\in \{1,2,3,...\},
    \label{eq:affinit_matrix}\vspace{-2mm}
\end{equation}\endgroup
where $L=\sum_{t}(D_t-A_t)=\sum_{t}L_t$. Consequently for each $t$, we have $L_tE = (D_t-A_t)E$, such that $L_t$ has each row sum to zero. Therefore, we can draw the conclusion that $L$ is a Laplacian matrix and that propagation via our proposed SGPN module represents a standard diffusion process on graphs.
 
\section{More Implementation Details}\label{sec:settings}
We provide more details about our implementation of the experiments presented in the paper, including the computational settings and run-times (Section~\ref{sec:compute}), re-organization of arbitrarily-structured data (Section~\ref{sec:top}), the detailed training and testing procedures for point clouds (Section~\ref{sec:point}), as well as the network architectures (Section~\ref{sec:CNNs}).

\subsection{Computational Settings and Run-times}\label{sec:compute}
\textbf{Computation.} We implement all our experiments on servers equipped with four 16GB Titan V100 GPUs.
All scripts are implemented with the PyTorch framework\footnote{https://pytorch.org/} and will be made available to the public upon publication. \\

\textbf{Training.} 
Specifically, we utilize the data parallel module of PyTorch to accelerate training and testing, but we \textbf{do not} use any multi-GPU batch normalization techniques even for the Deeplab-related models, in order to maintain a basic setting so that we can focus on the capability of, and the improvement that can be brought by SGPN.
We train all hybrid models, \textit{i.e}., CNN block+SGPN jointly with weights initialized for the CNN from ImageNet.
The only exception to this is the experiment with NLNN~\cite{Wang_nonlocalCVPR2018}, where better performance is achieved when the CNN is initialized with weights from the baseline backbone segmentation network (see Table 1 in the paper).
We use most of the hyper-parameter values that were originally proposed to train the baseline networks (\eg, learning rate, weight decay, etc.). In addition, we use the SGD and Adam solvers for the DRN and Deeplab-related networks, respectively. \\

\textbf{Run-times.}
The inference time of one image ($1024\times 2048$) from the Cityscapes dataset is 0.2 and 0.5 second for DRN-based SGPN on pixels (Section 5.3) and superpixels (Section 5.4), respectively (on a V100 Nvidia GPU).
It takes approximately two minutes to carry out inference with our hybrid networks (\eg, the CNN block and the SGPN on the entire point cloud in Fig. \ref{fig:gpus}), for all the images and points in the validation set.
%
\begin{figure*}[t]
\centering
    \begin{minipage}[t]{0.45\linewidth}
        \includegraphics[width=0.99\linewidth]{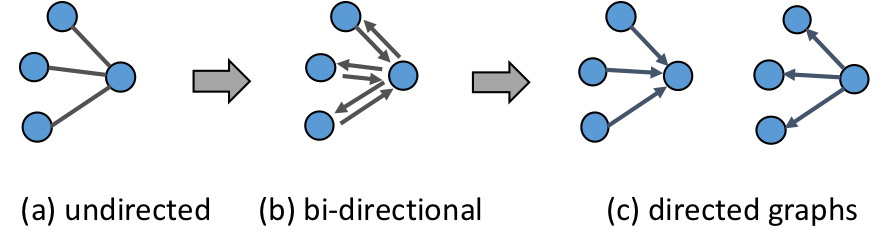}
        \caption{\small We construct the DAGs by: (a) establishing an undirected graph; (b) taking (a) as bi-directional graph, and (c) decomposing it into several directed graphs.}
        \label{fig:01}
    \end{minipage}
        \hfill%
    \begin{minipage}[t]{0.45\linewidth}
        \includegraphics[width=0.99\linewidth]{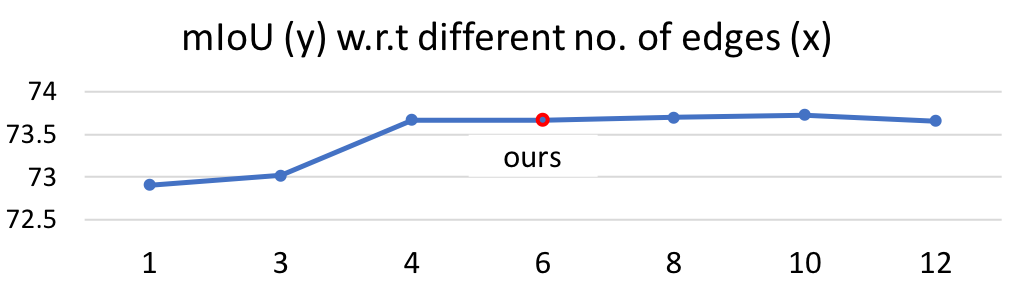}
        \caption{\small Perturbing the number of edges (axis-x) for each vertex in the constructed DAGs. Performance is shown using the \textit{+PF+PG/TG} model (see Tabel.2 in the main paper).}
        \label{fig:02}
    \end{minipage} 
\end{figure*}

\subsection{Discussion about DAGs}\label{sec:top}
\paragraph{DAG Construction.} For constructing DAGs, we first search for the K-nearest neighbors of each point (please refer the next paragraph for more details), then re-organize them into different groups for parallel computation (see Section 3.1 in the paper).
Re-organization of vertices in a DAG can be easily carried out via a modified topological sorting algorithm~\ref{algo:toposort}. We use $p$ to denote the index of groups, as described in Section 3.1 in the paper and in Eq.~\eqref{eq:group}. 
Specifically, finding the neighbors of superpixels can be conducted by searching for the adjacent superpixels that have shared boundaries. Note that since the superpixels are irregular, it can sometimes happen that two superpixel's center overlap. This will cause there to be directed cycles in the directed graphs. For such situations, we assign random small offsets to such overlapping centers to avoid this degnerate case.
For a point cloud, we fix the number of nearest neighbor to $K=6$ for all experiments presented in the paper.
%
After constructing the DAGs, the linear propagation in Eq.~\eqref{eq:group} is conducted along a constructed DAG, \eg, from the first group to the last one in the bottom-to-top DAG, as shown in Fig.~\ref{fig:topogroup}.

\paragraph{Structures of the DAGs: a symmetric affinity matrix.}
A undirected graph structure is preferable to eliminate any structural biases. 
Although we propose the directed graphs, we do ensure that the linear propagation on DAGs equal to a symmetric affinity. 
Ideally, an undirected graph should be established and a single weight is attached to each edge.
We ensure the symmetry in two aspects: 
(a). \textit{The graph is bi-directional}: we first construct an undirected graph by finding the nearest neighbors for each element while maintaining an symmetric adjacency matrix (Fig.~\ref{fig:01} (a)). Then, we take it as a bi-directional graph and decompose it into several directed graphs w.r.t different directions (Fig.~\ref{fig:01} (b)(c)).
(b). \textit{The weight is symmetric}: Each pair of bi-directional links share a single weight since the kernels are symmetric, i.e., $\kappa (x_i, x_j) = \kappa (x_j, x_i)$. See Section 3.2 in the paper.
In addition, we consider all possible directions (i.e., 4 for 2D and 6 for 3D domain) to ensure the bi-directional graph is fully utilized by the propagation module. 
\begin{algorithm}
\centering
\caption{Re-ordering of vertices in a DAG}\label{algo:toposort}
\begin{algorithmic}[1]
\State $G$: the DAG to be re-ordered.
\State $L$: an empty list that will contain the groups for all the ``time-steps''.
\State $S$: the parent nodes without any incoming edges.
\State $p=1$ denotes the index of the group, starting from 1.
\State \emph{top}:
\If {$L$ collected all the vertices} \Return
\EndIf
\State \emph{loop}:
\State remove S from G
\State $L(p) \gets S$ (Note that $L(p)$ is a sub-list of $L$.)
\For{{\textit{each node $m$ with edges $e$ coming from $S$}}}
    \State \textit{remove $e$ from $G$}
\EndFor
\State update $S$ based on $G$, $p=p+1$
\State \textbf{goto} \emph{top}.
\end{algorithmic}
\end{algorithm}
\vspace{-3mm}
\paragraph{Different choices of established DAGs.}
We conduct ablation studies for different choices of DAGs in two aspects: (a) changing the directions on the construction of a DAG, and (b) perturbing the number of connections for each nodes.
Since we proposed a factorized edge representation, our network is quite flexible to be directly tested on the above settings, without re-training.
We show that the performance is not sensitive to the variants of DAGs, which also validates that it is similar to an undirected graph.
 
(a). \textit{Changing the directions.} On the RueMonge2014 dataset, we rotate the point cloud globally in the XY plane with an angle of 20 degrees, and re-establish the DAGs. The neighbors remain the same for each point, but the directions for propagation are different (i.e., rotating the axis-X for propagation with 20 degrees). With the \textit{+PF+PG/TG} model, we obtain similar mIoU (73.32\%) as the original one (73.66\%).

(b). \textit{Perturbing the number of edges.} This study is introduced in l-(201-203) in the supplementary material. In detail, we perturb the number of neighbors $K$ for each vertex in a DAG from 1 to 12. We obtain similar performance especially when $K>3$. See Fig.~\ref{fig:02}. 

\begin{figure}[t]
    \centering
    \includegraphics[width=0.99\linewidth]{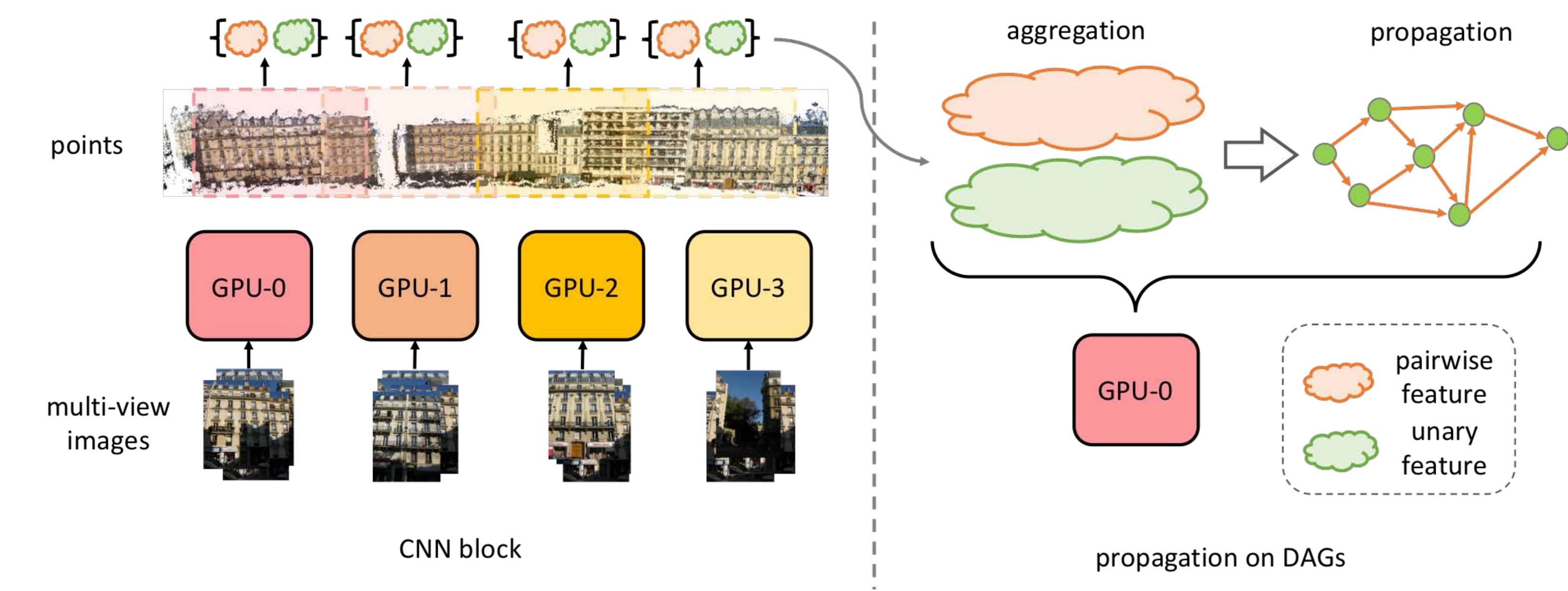}
    \caption{Inference of multi-view images + a point cloud (we use the cloud-shaped blobs to present the the aggregated features for the 3D points).
    CNN-block: it infers all the multi-view images in parallel, and aggregates the unary and pairwise features on each GPU independently. Note that points between images and across GPUs are highly overlapped.
    Aggregation (across GPUs): the unary and pairwise features from each GPUs are re-aggregated across GPUs in order to avoid duplication.
    Propagation on DAGs: The SPN module is run on a single GPU.
    }
    \label{fig:gpus}
\end{figure}

\subsection{Processing of point clouds}\label{sec:point}
The training and testing procedures for point clouds in this work are slightly different from that of superpixels. We alluded to them at a high-level in the paper, but here, we specify them in more detail and provide more intuition.

\paragraph{How to augment for propagation on graphs?}
Augmenting 3D points is usually conducted by extending data augmentation techniques used to augment image patches to 3D space~\cite{pointcnn,pointnet,pointnetplusplus}.
%
In our work, however, this is made even simpler by the fact that all the augmentations can be conducted simply on 2D space -- the cropped image patches, while training a CNN with images.
Since we maintain an index map of the same size as an input image, all augmentations conducted on images can be simultaneously applied to their index maps.
The augmented points are associated with these augmented index maps.

\paragraph{Propagation on a global scope.}
During training, we utilize multiple GPUs where each one takes a mini-batch of image patches, aggregates image features according to the indices of the subset of \emph{local} points corresponding to those patches, and conducts propagation and MLP before computing the loss w.r.t the ground-truth labels.
However, the testing procedure for a point cloud is different from its training procedure, in the sense that while inference of individual images is distributed across different GPUs, the propagation is conducted \emph{globally} on the entire point cloud, by collecting information across the GPUs. 
In contrast, if we were to test/process each image along with its corresponding set of points individually, propagation would be limited to only a local set of points.
%
To verify this, we test each image individually, and for every 3D point aggregate DRN results of all pixels across all images that map to it, as opposed to processing the images collectively and propagating information globally in the point cloud (\emph{image to points} in Table 2 in the paper), and the mean IoU drops from $69.16$ to $65.50$.
Clearly, aggregating and propagating on a global level is of great importance for labeling points.
Fortunately, the flexibility of our algorithm brought about by the factorization of the edge representation (Section 3.3), enables the algorithm to be implemented in a highly flexible, yet efficient manner as shown in Fig.~\ref{fig:gpus}.

In detail, we divide the inference phase into the stages of (a) inference of all the testing images, which can be done concurrently for each image on multiple GPUs (the CNN block in Fig.~\ref{fig:gpus}), and (b) global aggregation of all the unary and pairwise features from all images and mapping them to the entire point cloud, which can be done on a single GPU (the propagation on DAGs in Fig.~\ref{fig:gpus}).
We show our implementation in Fig.~\ref{fig:gpus}, where the CNN block is configured to perform inference on the images in parallel (\eg., on 4 GPUs), and the propagation (right) is done globally, after collecting the feature of all the images from multiple GPUs on to a single GPU, and aggregating them for the entire point cloud.

\begin{figure}[t]
    \centering
    \includegraphics[width=0.95\linewidth]{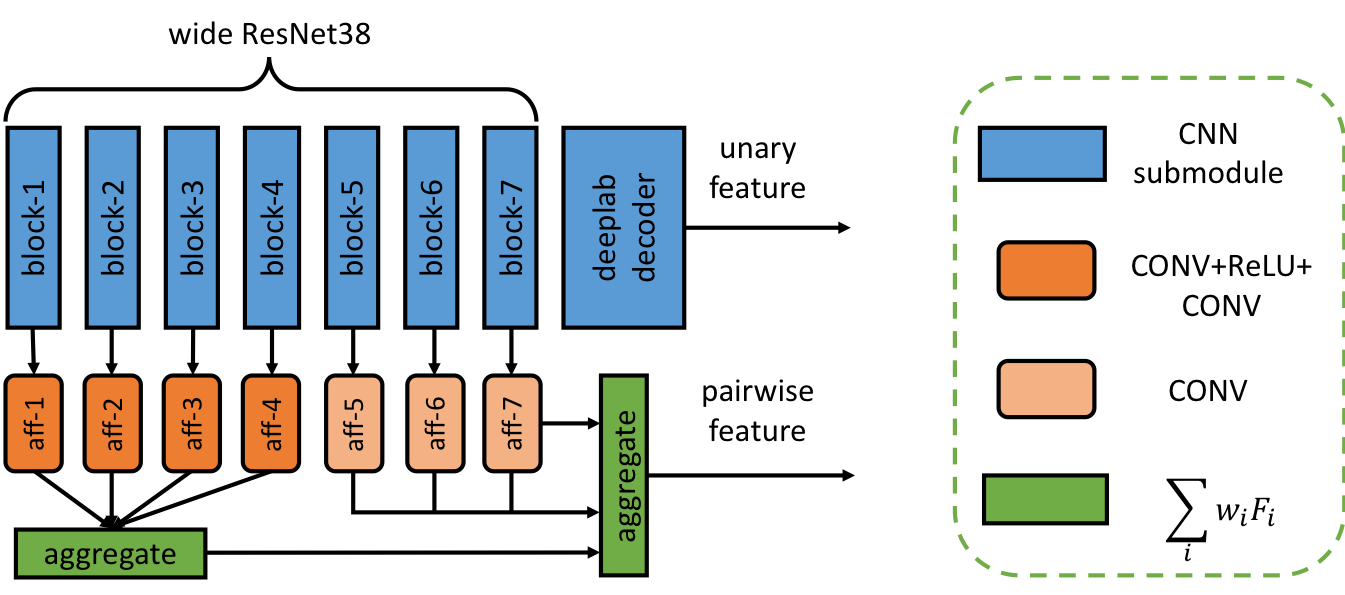}
    \caption{CNN block for the Deeplab network-based~\cite{deeplabv3plus2018} SGPN.}
    \label{fig:cnn}
\end{figure}

\subsection{Network Architecture of the CNN Block}\label{sec:CNNs}
Here we show the architecture of a Deeplab-based CNN block, as discussed in Section 5.3 in the paper, in Fig.~\ref{fig:cnn}.
This is also a detailed version of the first module in Figure 4 of the paper.
Specifically, the CNN sub-module contains a series of residual blocks~\cite{widerres}.
The aggregation of different feature levels of the CNN is conducted with $\sum_i w_iF_i$, where $w_i$ is a learnable, scalar parameter for level $i$ initialized to $1$, and $F_i$ denotes the feature map from level $i$.

\section{More Experimental Results}\label{sec:exp}

\subsection{Semantic Segmentation}
\begin{figure*}[h]
\centering
\begin{tabular}{c@{\hspace{0.01\linewidth}}c@{\hspace{0.01\linewidth}}c@{\hspace{0.01\linewidth}}c@{\hspace{0.01\linewidth}}c@{\hspace{0.01\linewidth}}c@{\hspace{0.01\linewidth}}c@{\hspace{0.01\linewidth}}c@{\hspace{0.01\linewidth}}}
		\includegraphics[height = .14\linewidth, width = .115\linewidth]{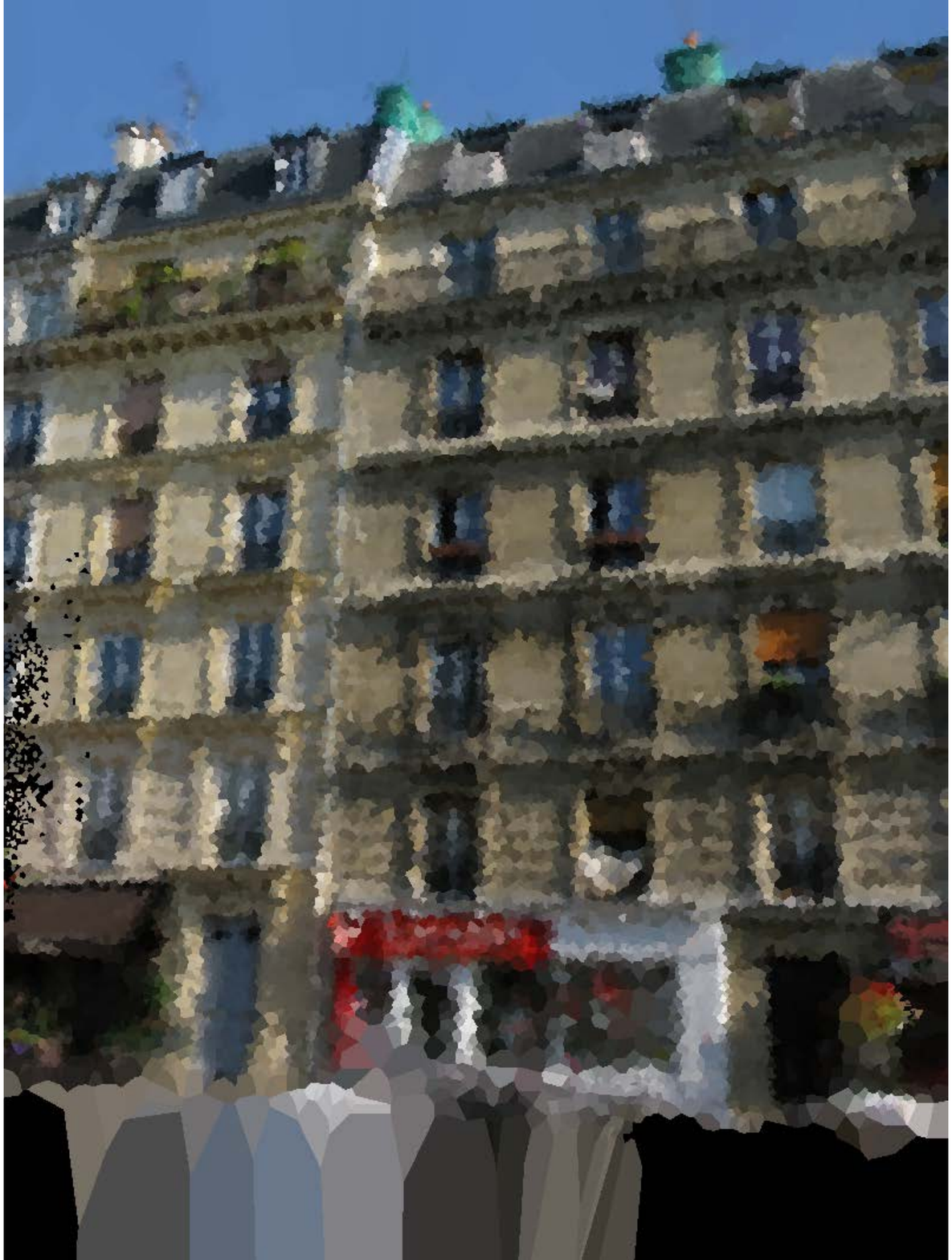} &
    	\includegraphics[height = .14\linewidth, width = .115\linewidth]{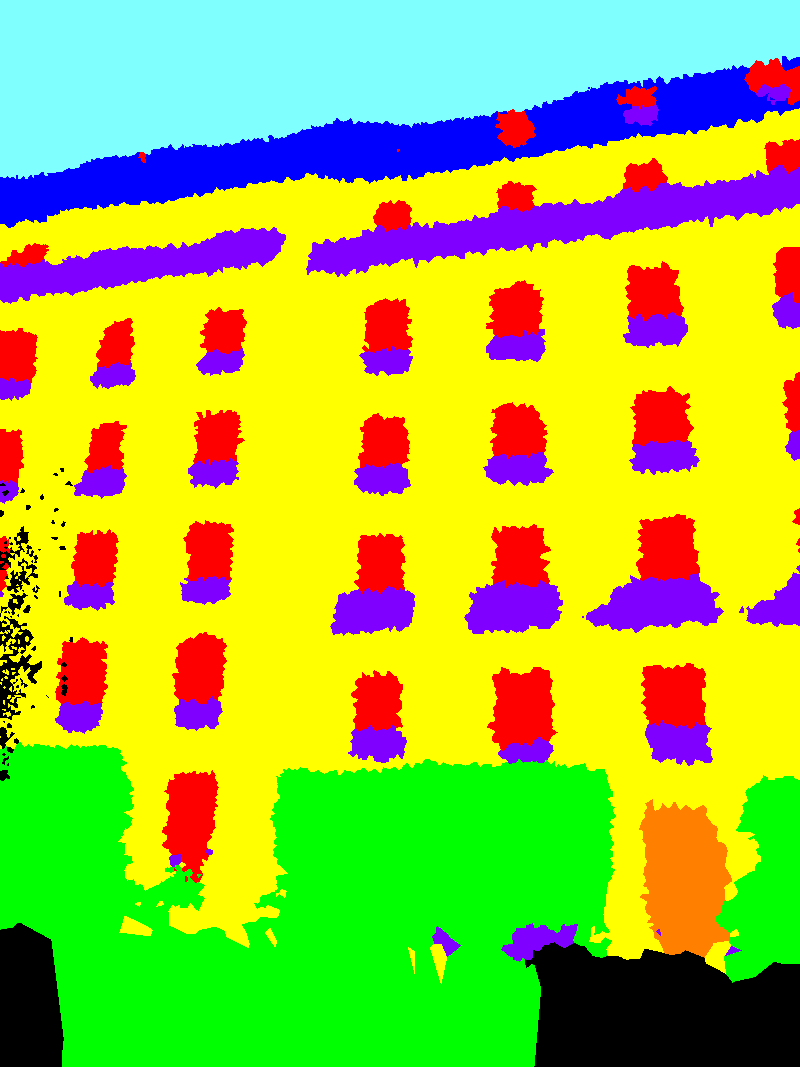} &
    	\includegraphics[height = .14\linewidth, width = .115\linewidth]{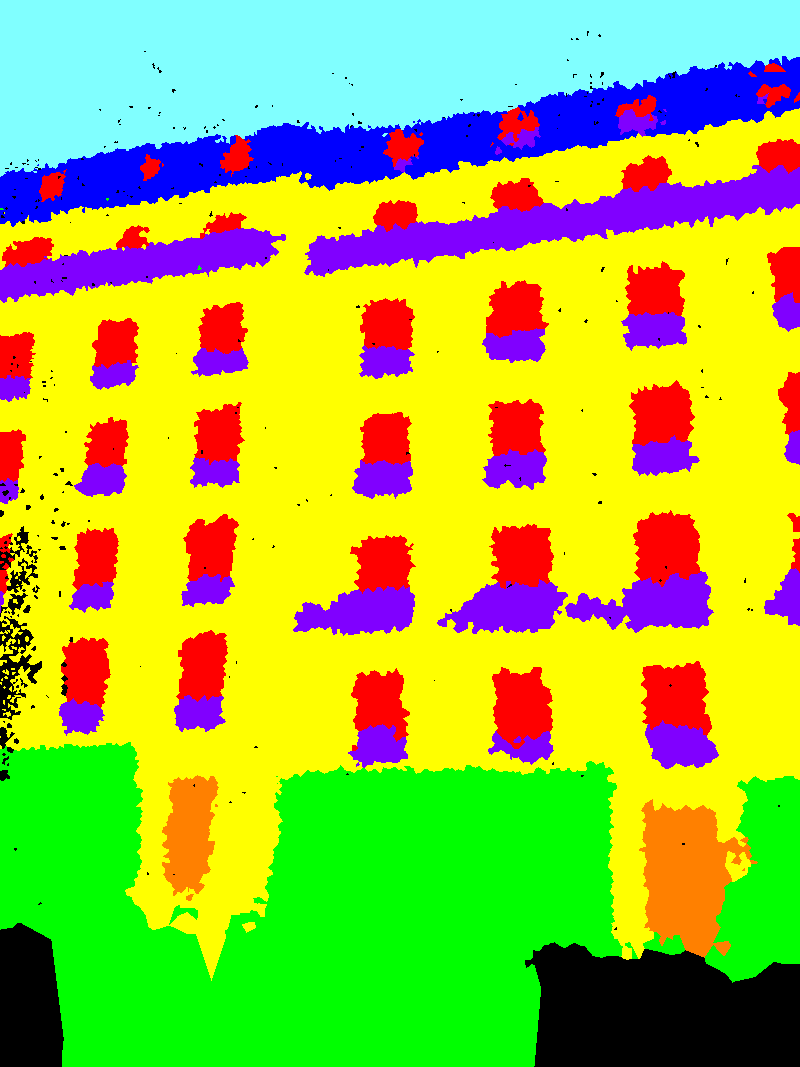} &
    	\includegraphics[height = .14\linewidth, width = .115\linewidth]{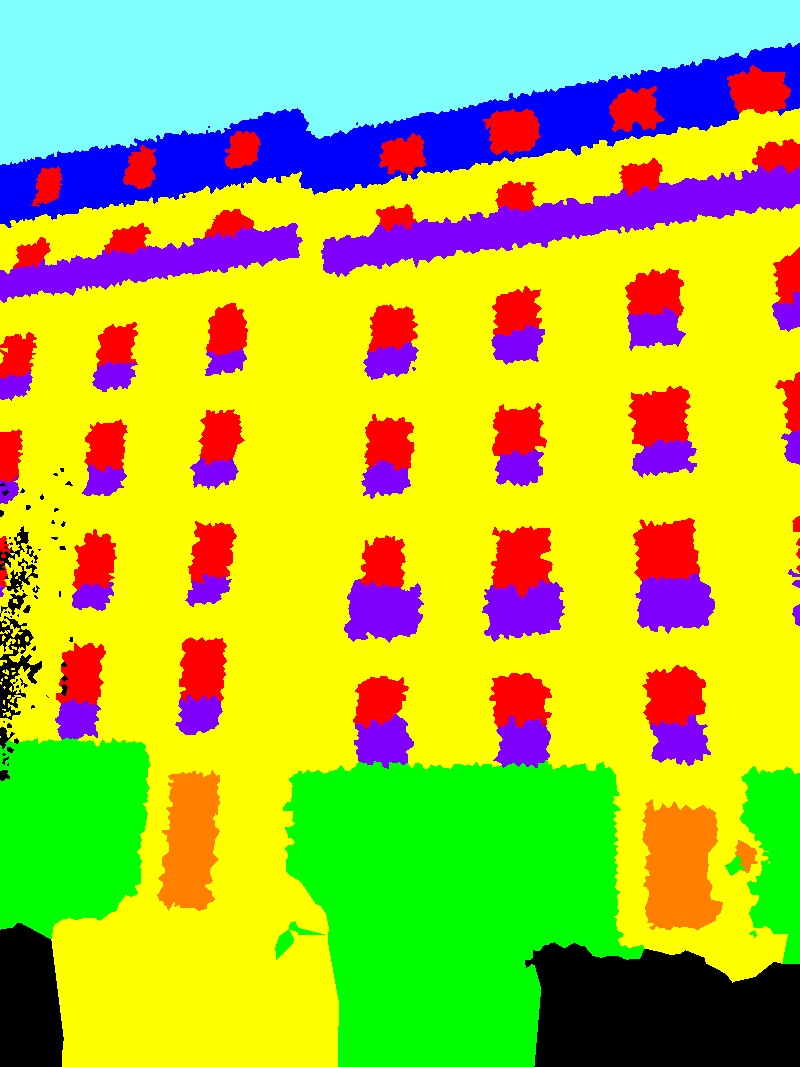} &
		\includegraphics[height = .14\linewidth, width = .115\linewidth]{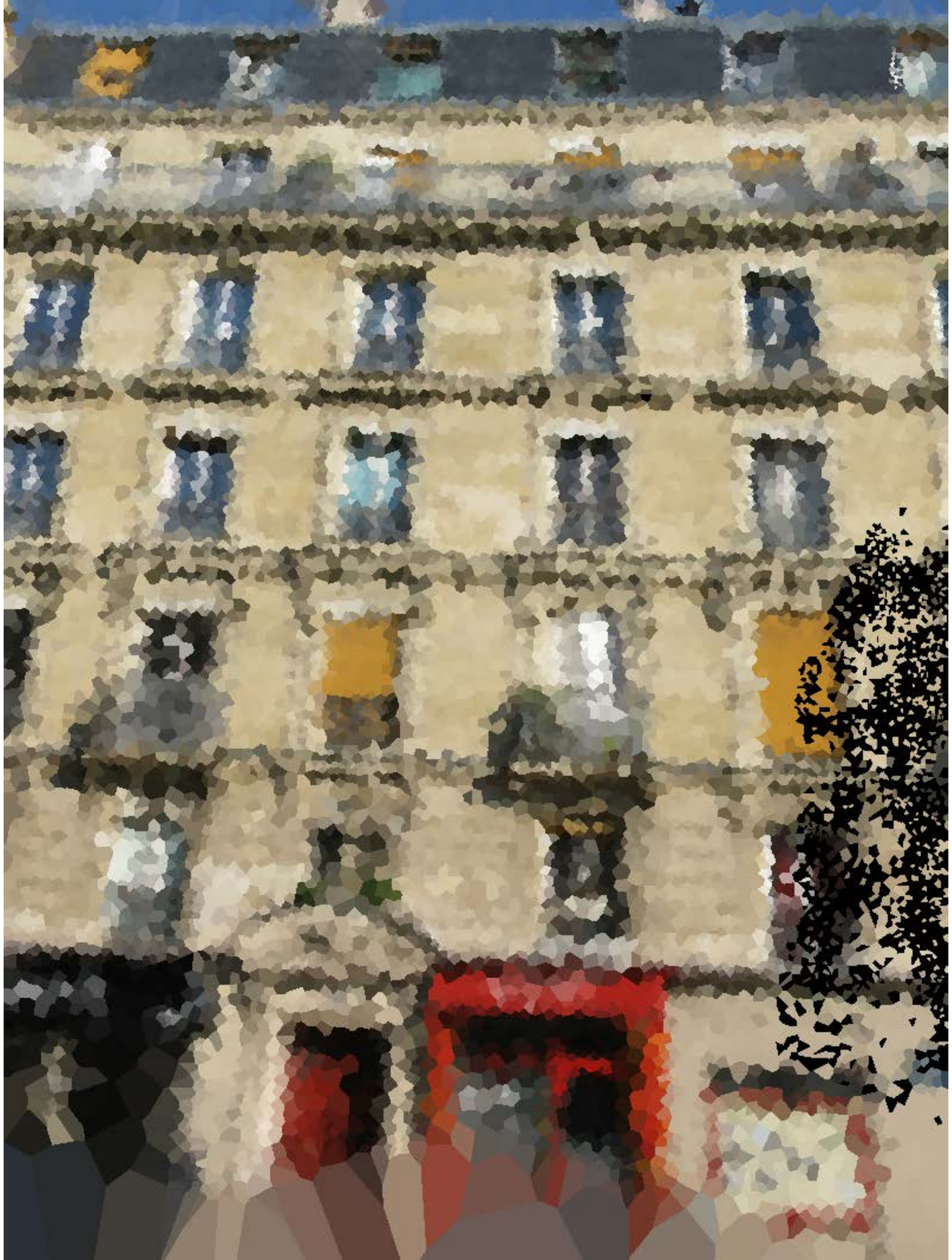} &
    	\includegraphics[height = .14\linewidth, width = .115\linewidth]{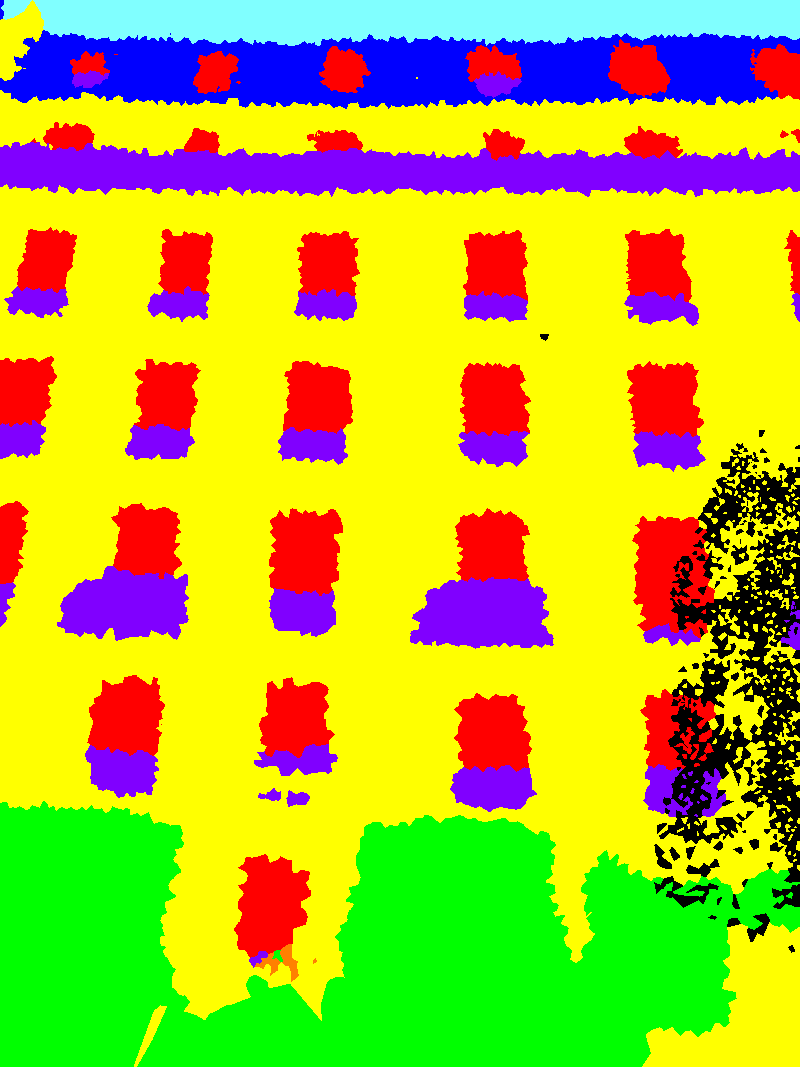} &
    	\includegraphics[height = .14\linewidth, width = .115\linewidth]{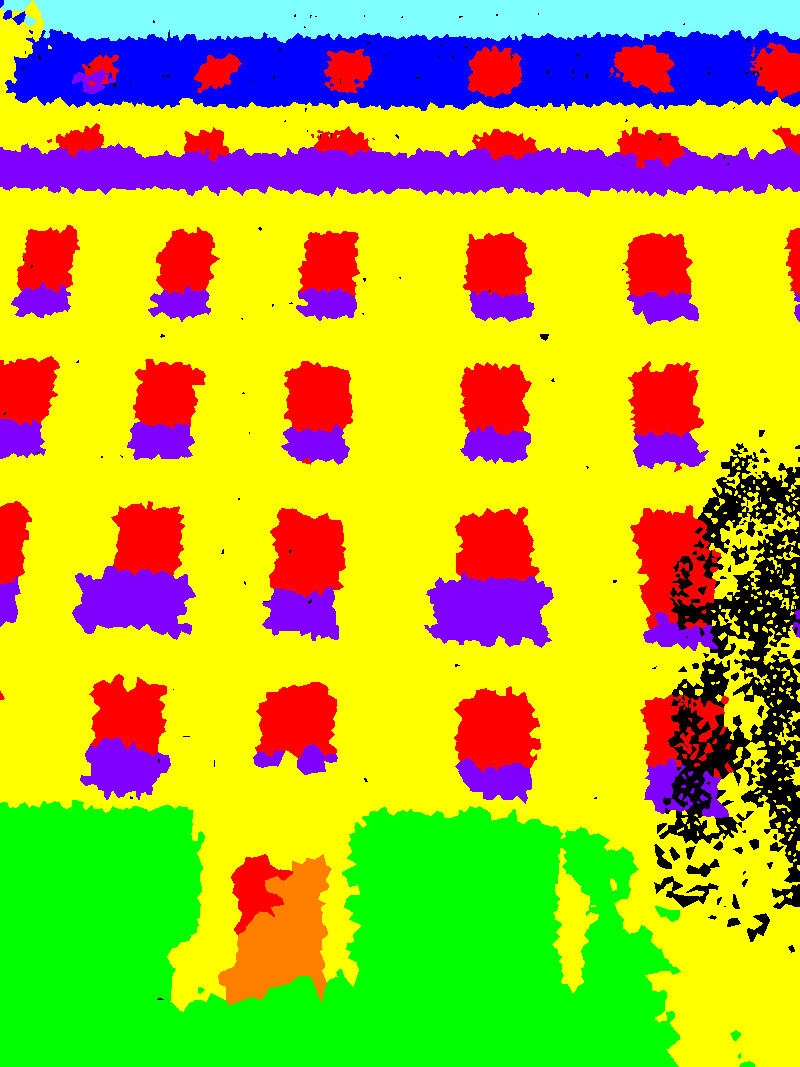} &
    	\includegraphics[height = .14\linewidth, width = .115\linewidth]{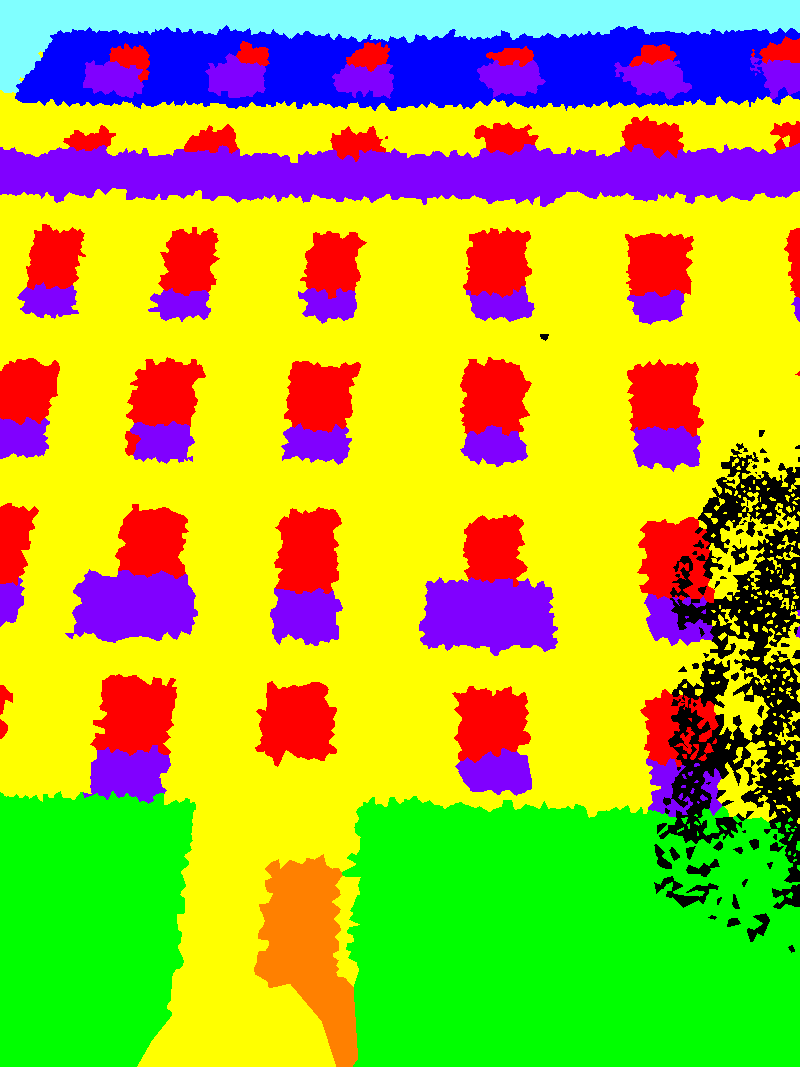} \\
		\includegraphics[height = .14\linewidth, width = .115\linewidth]{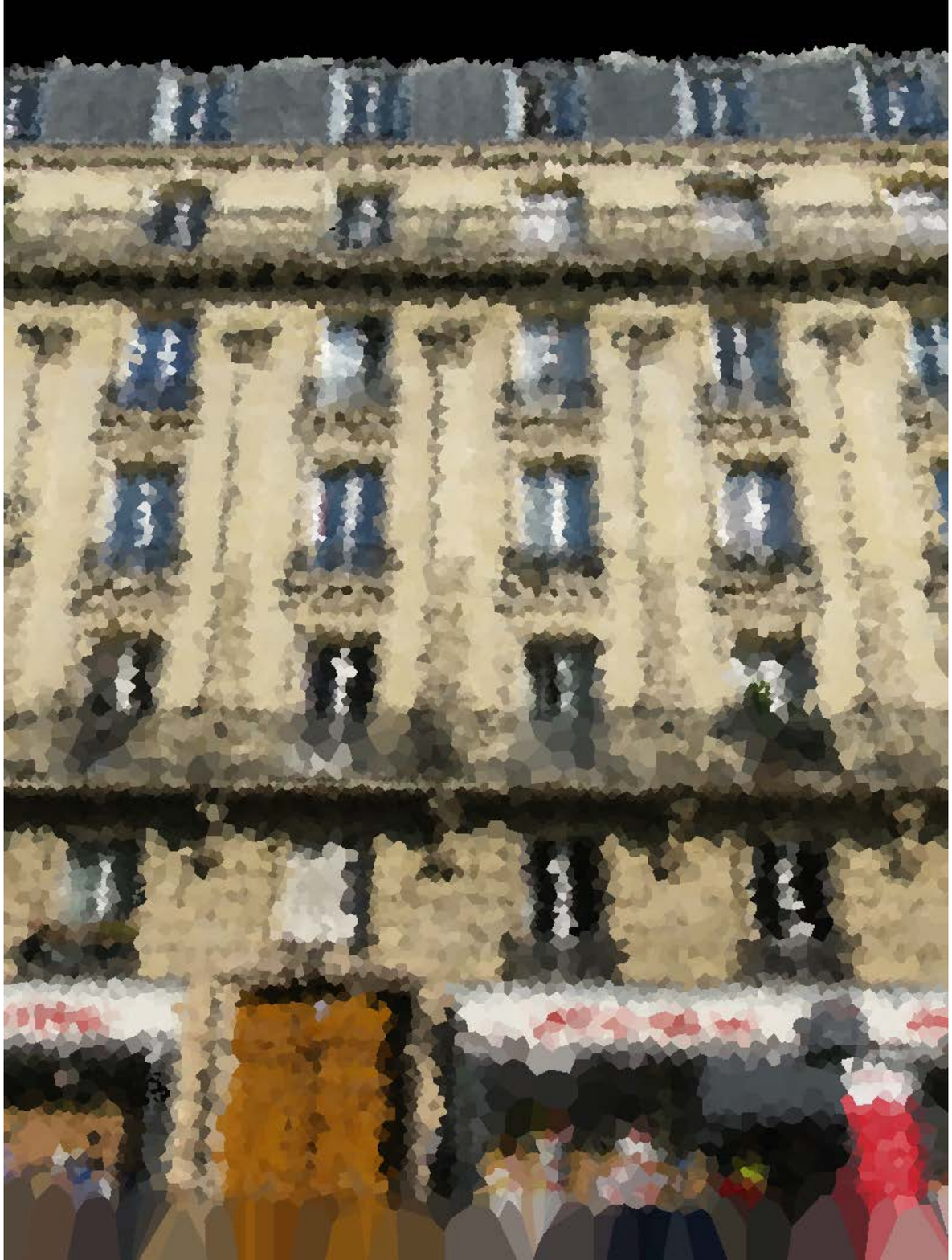} &
    	\includegraphics[height = .14\linewidth, width = .115\linewidth]{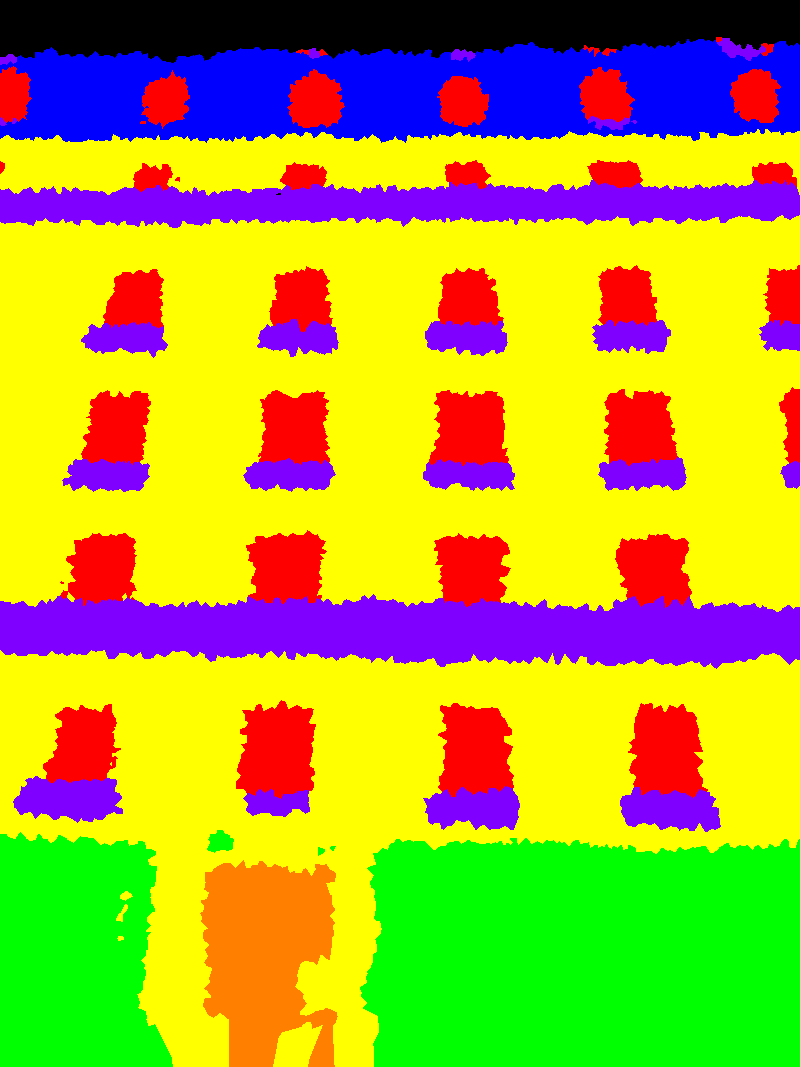} &
    	\includegraphics[height = .14\linewidth, width = .115\linewidth]{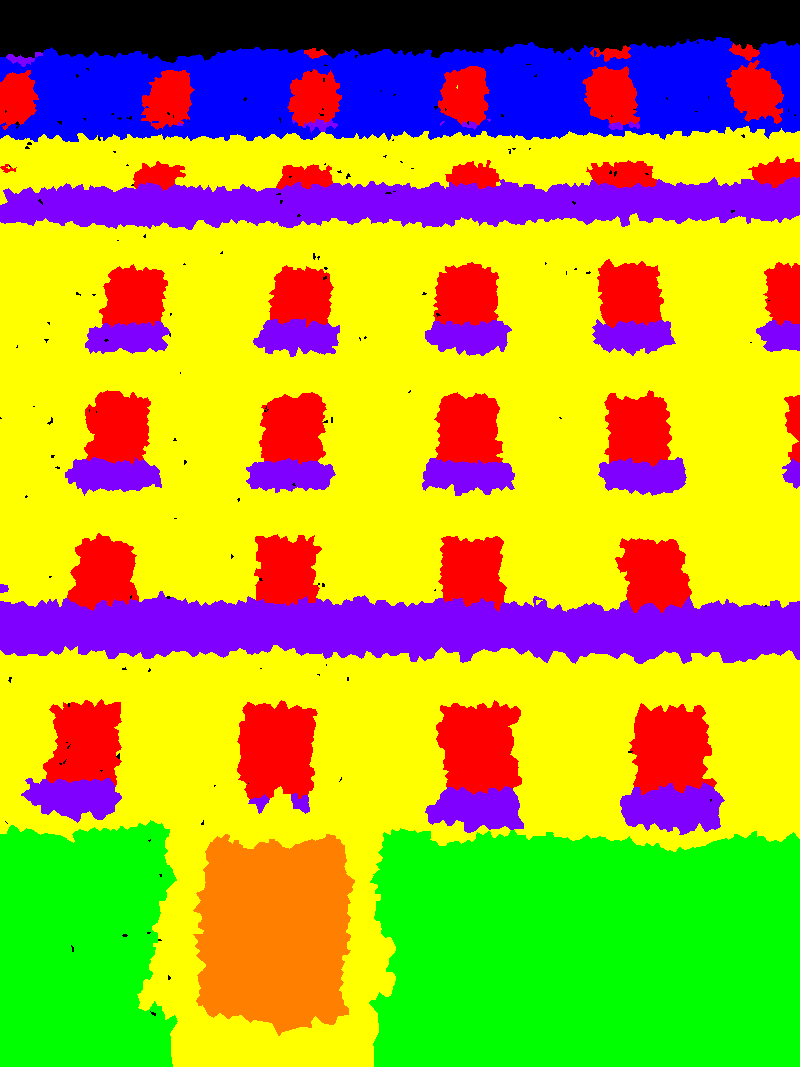} &
    	\includegraphics[height = .14\linewidth, width = .115\linewidth]{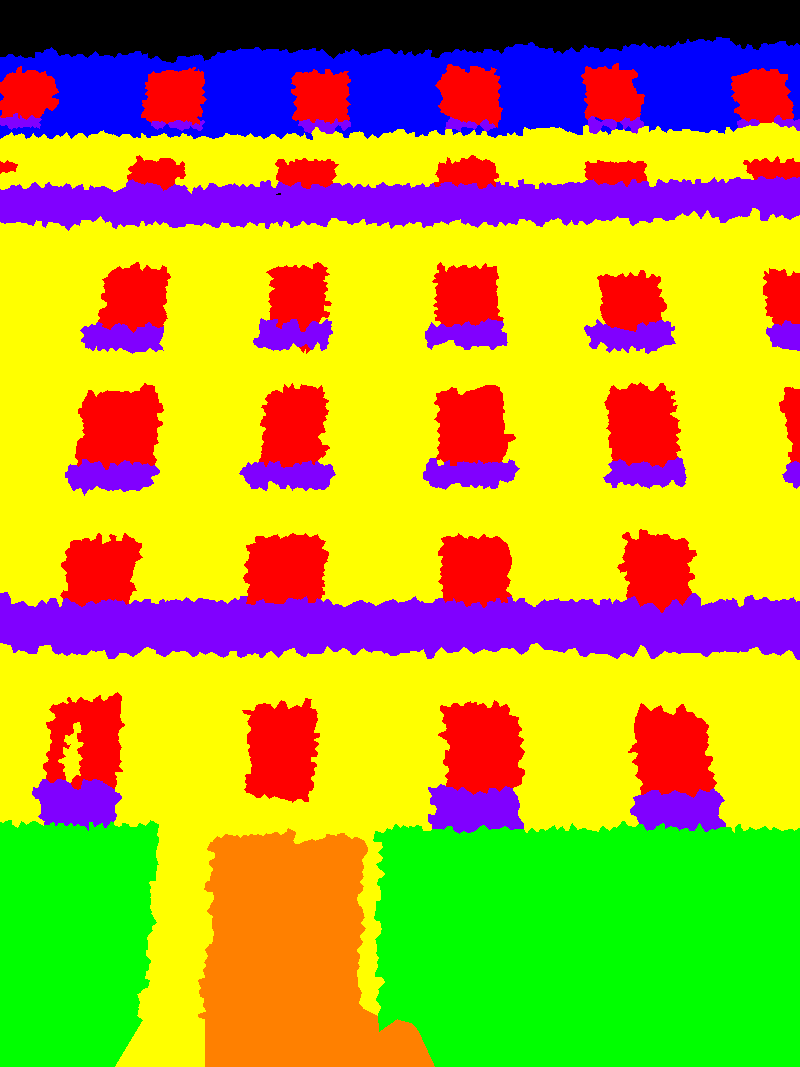} &
		\includegraphics[height = .14\linewidth, width = .115\linewidth]{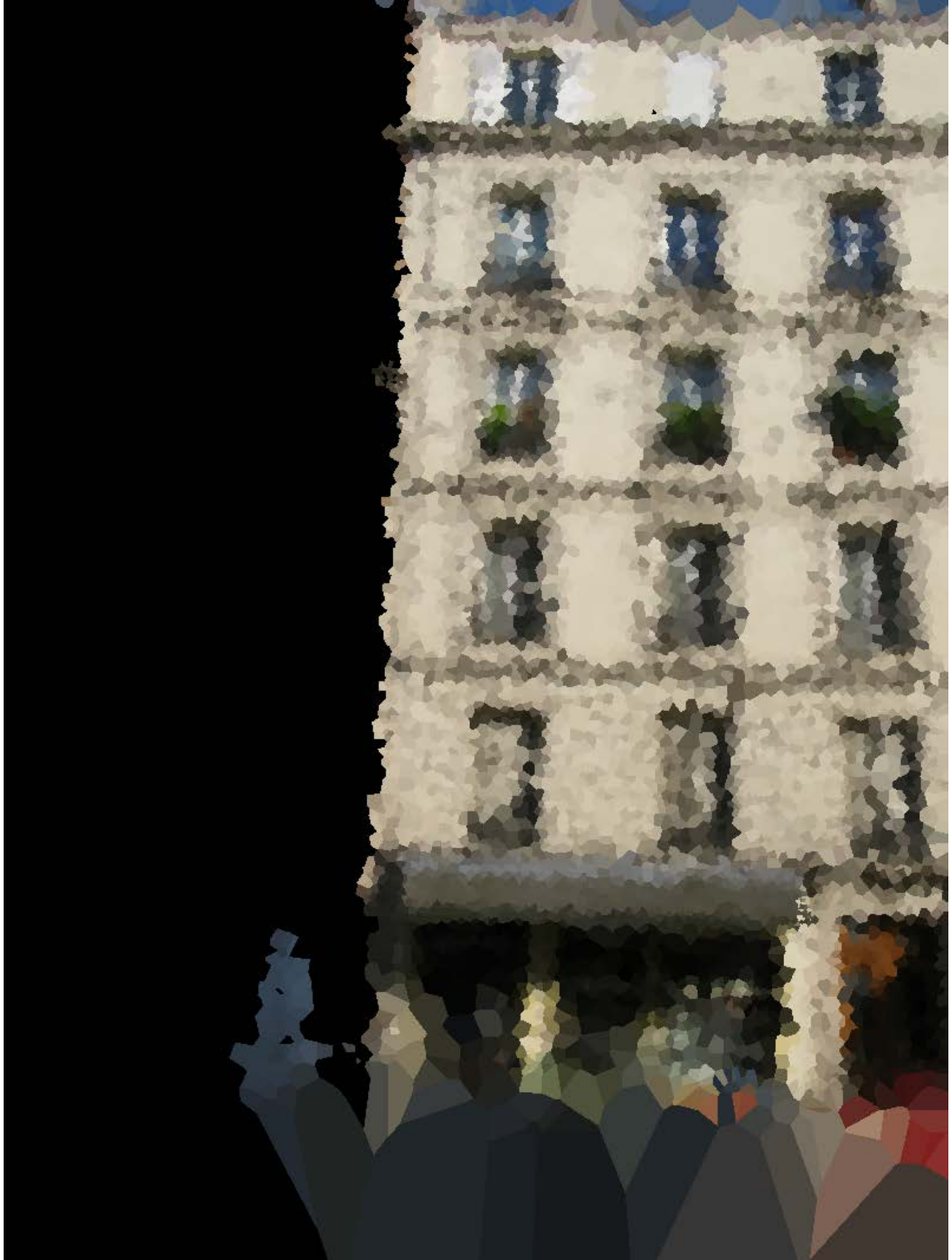} &
    	\includegraphics[height = .14\linewidth, width = .115\linewidth]{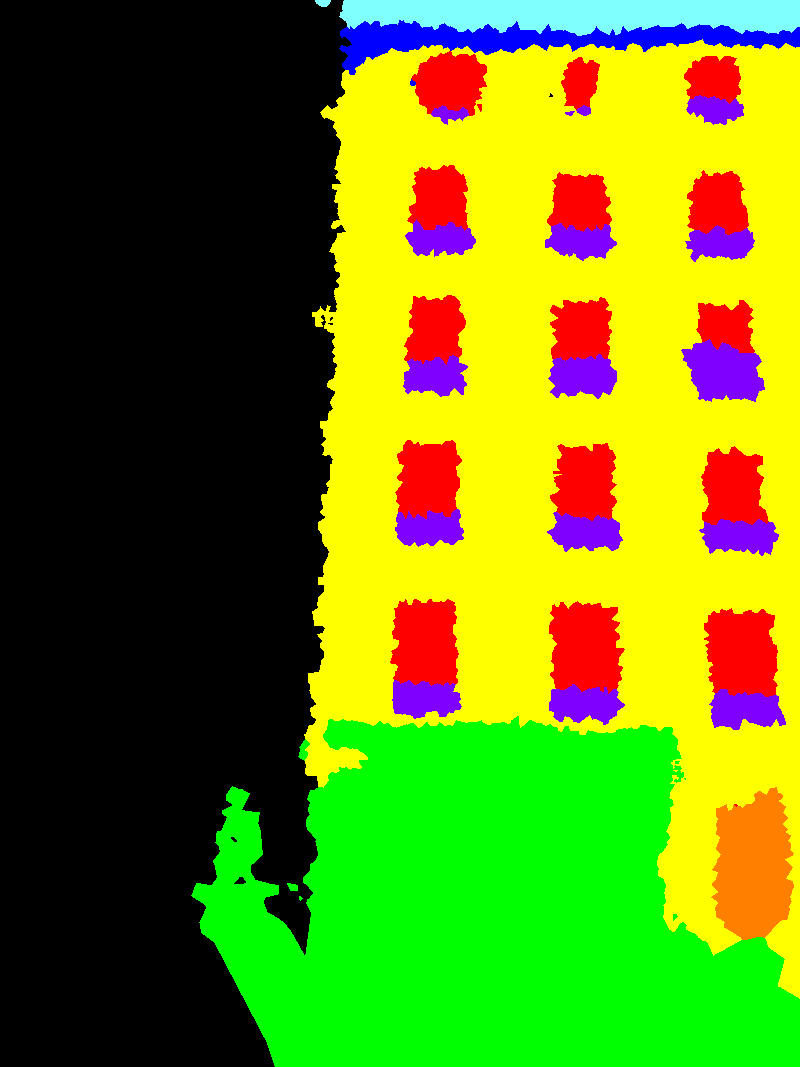} &
    	\includegraphics[height = .14\linewidth, width = .115\linewidth]{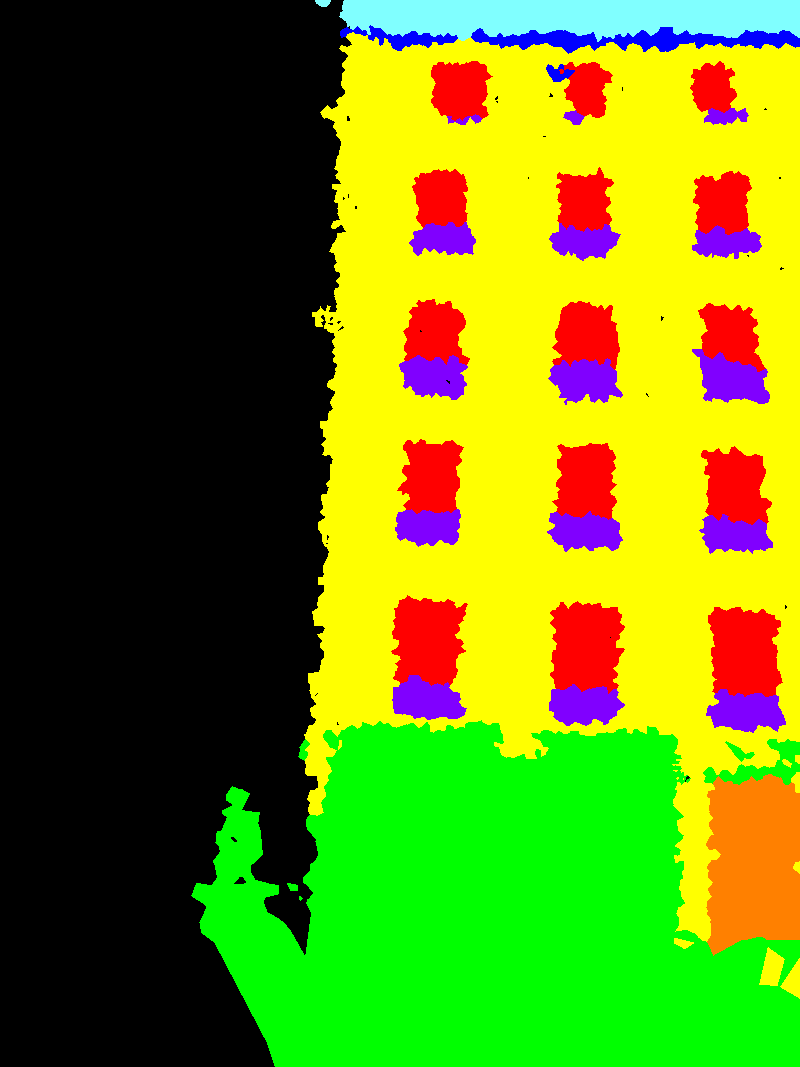} &
    	\includegraphics[height = .14\linewidth, width = .115\linewidth]{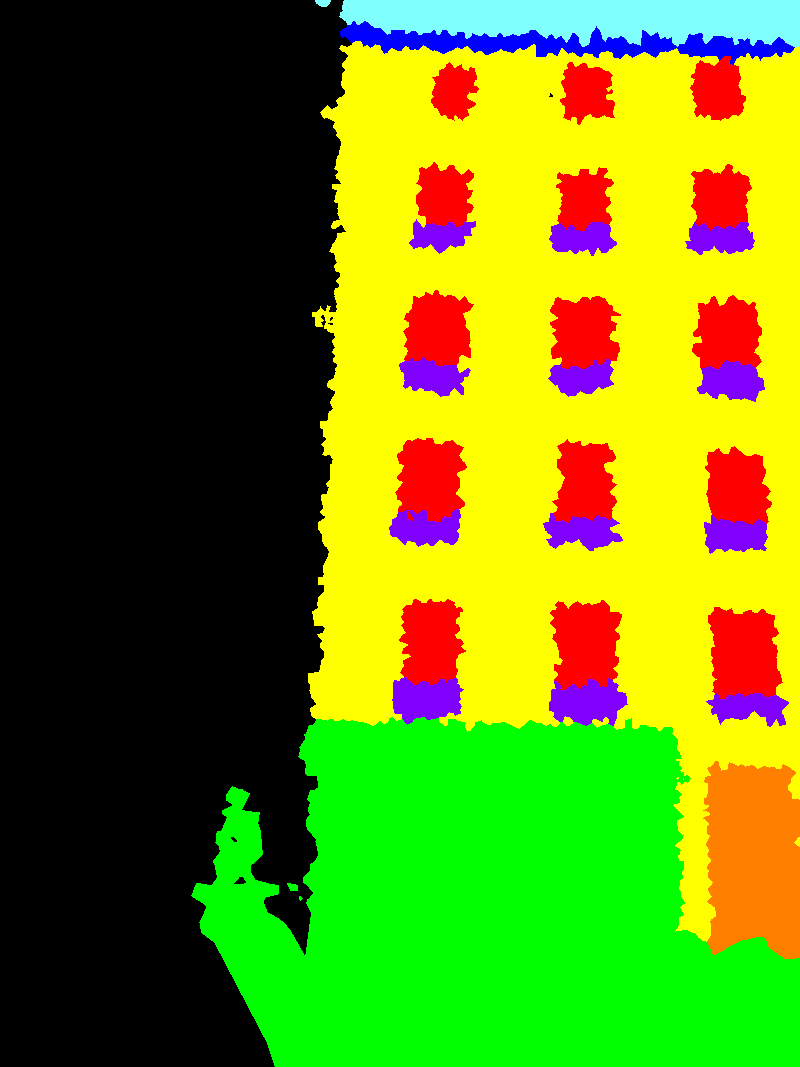}\\
		\includegraphics[height = .14\linewidth, width = .115\linewidth]{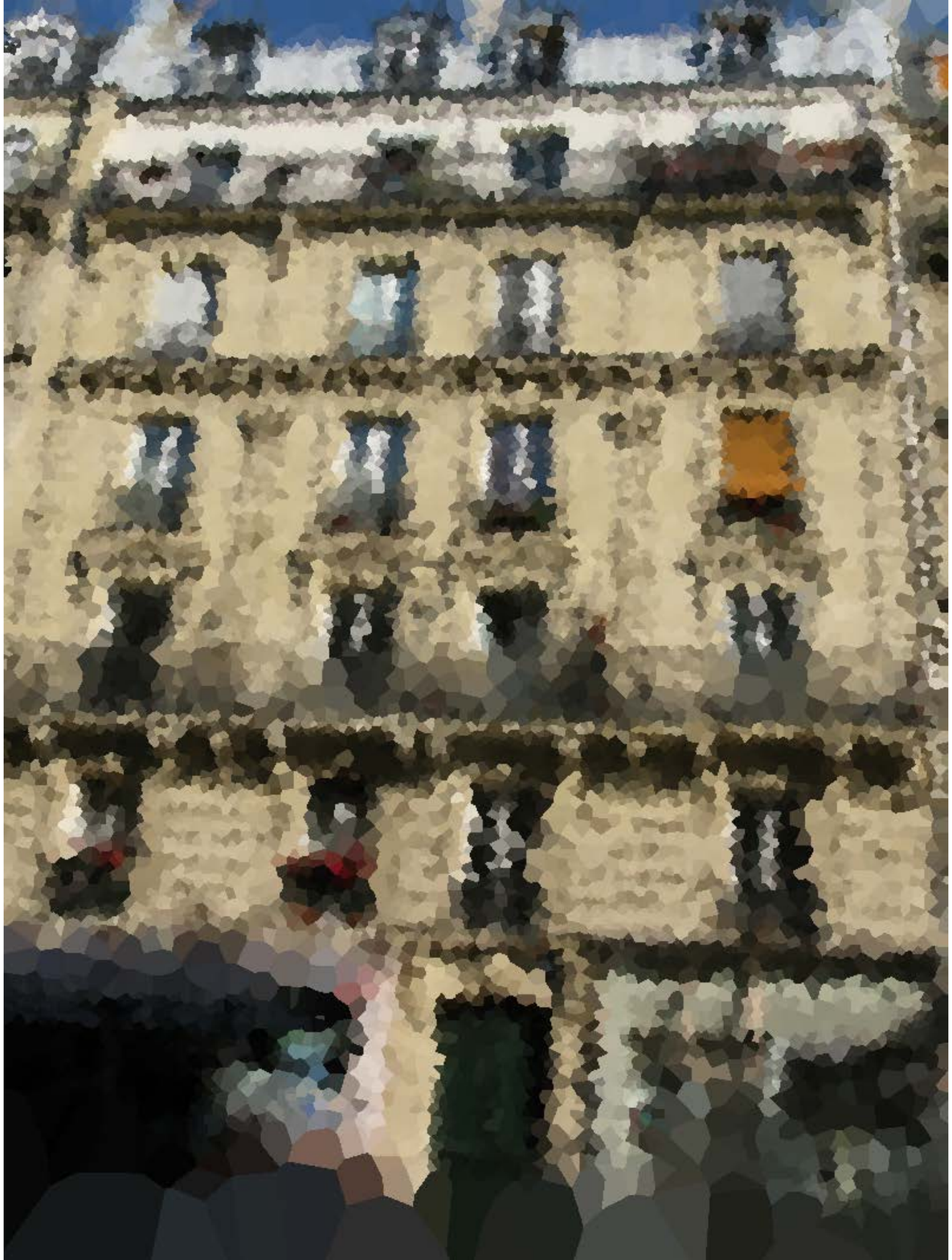} &
    	\includegraphics[height = .14\linewidth, width = .115\linewidth]{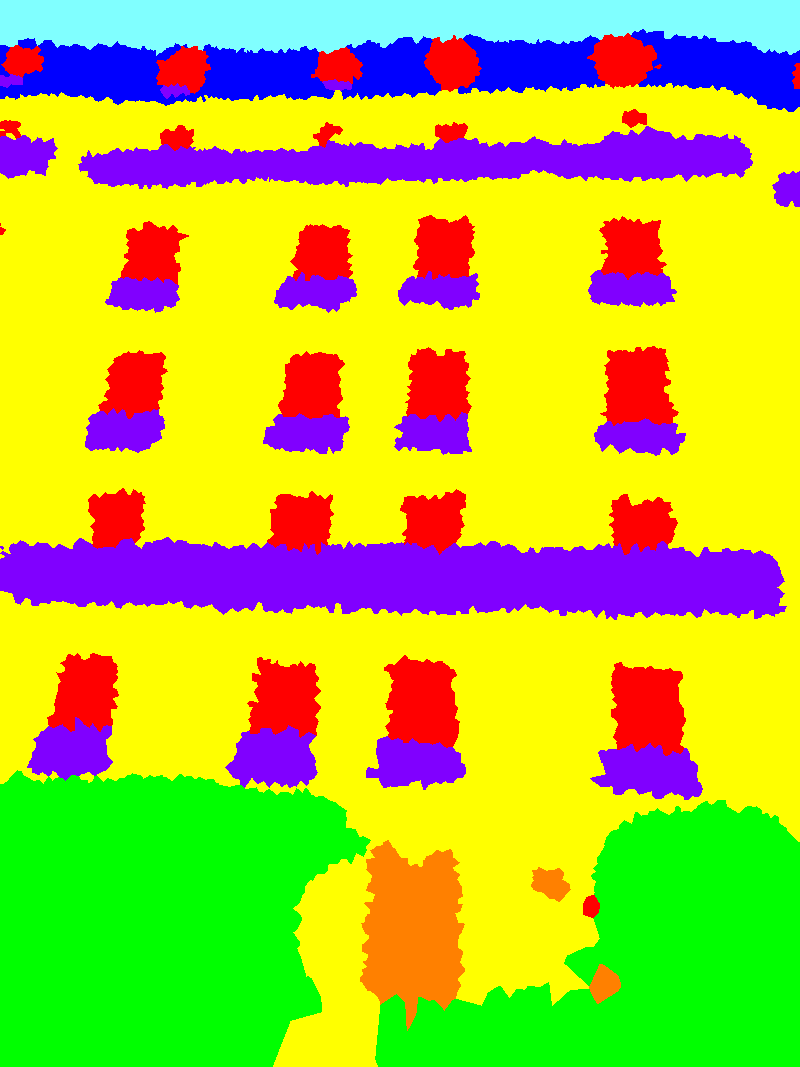} &
    	\includegraphics[height = .14\linewidth, width = .115\linewidth]{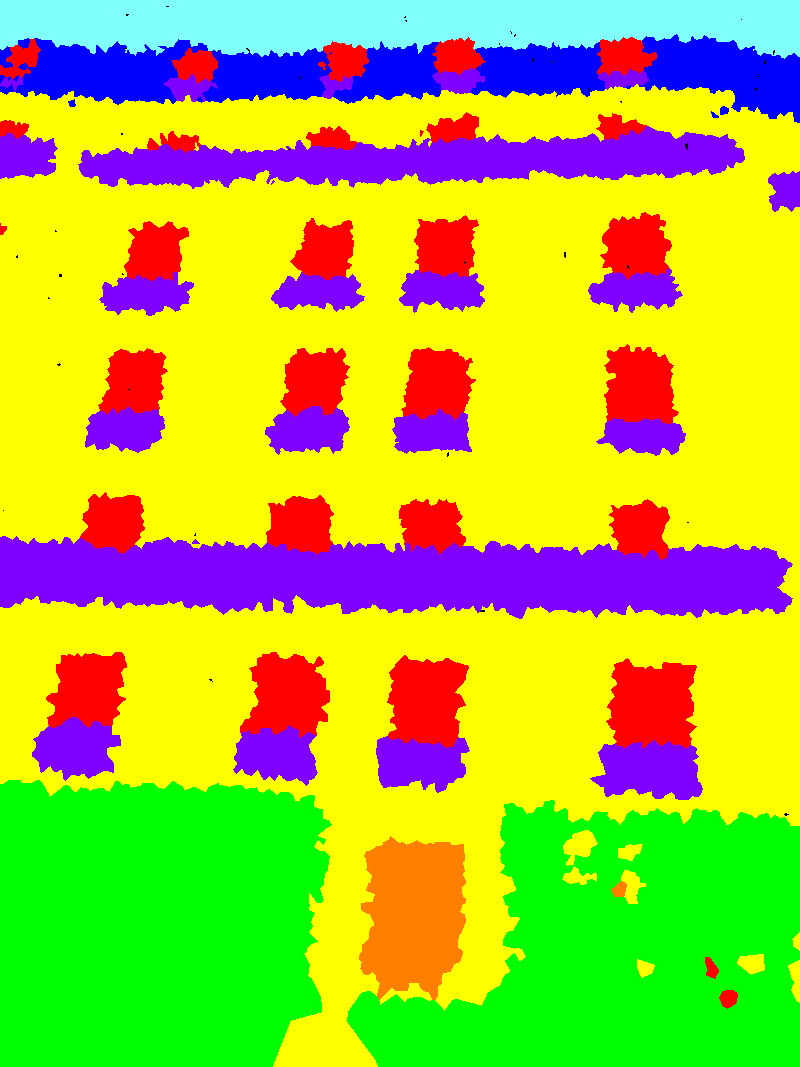} &
    	\includegraphics[height = .14\linewidth, width = .115\linewidth]{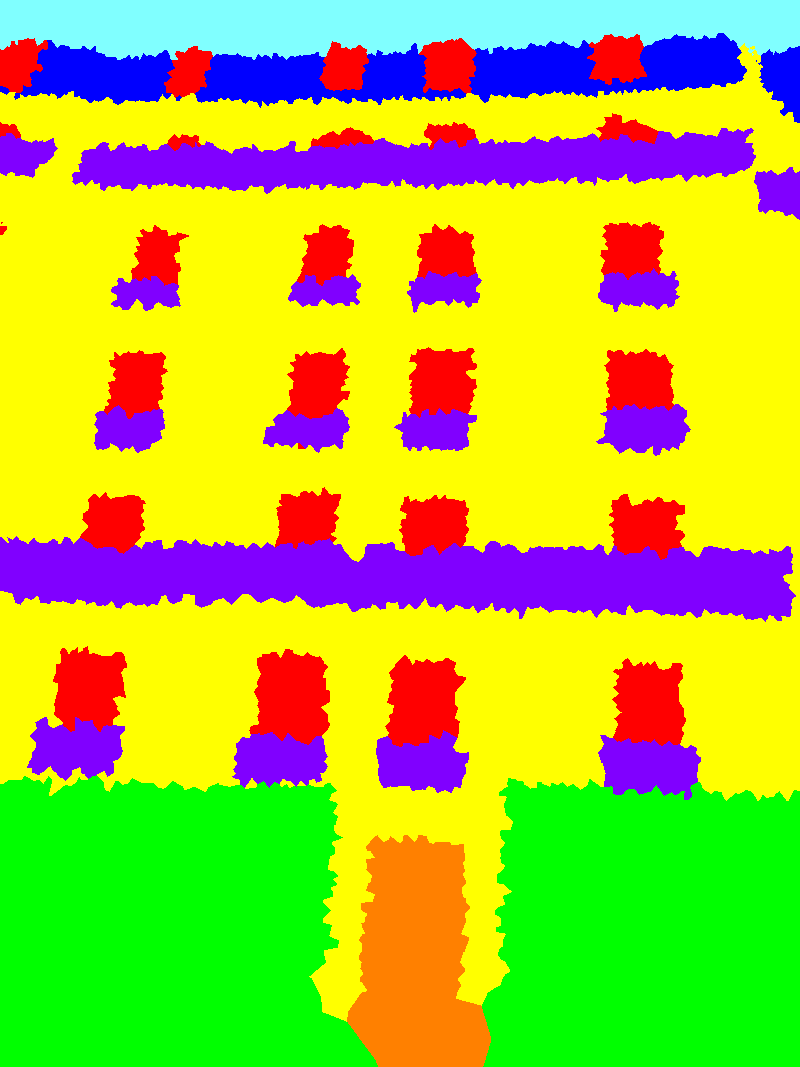} &
		\includegraphics[height = .14\linewidth, width = .115\linewidth]{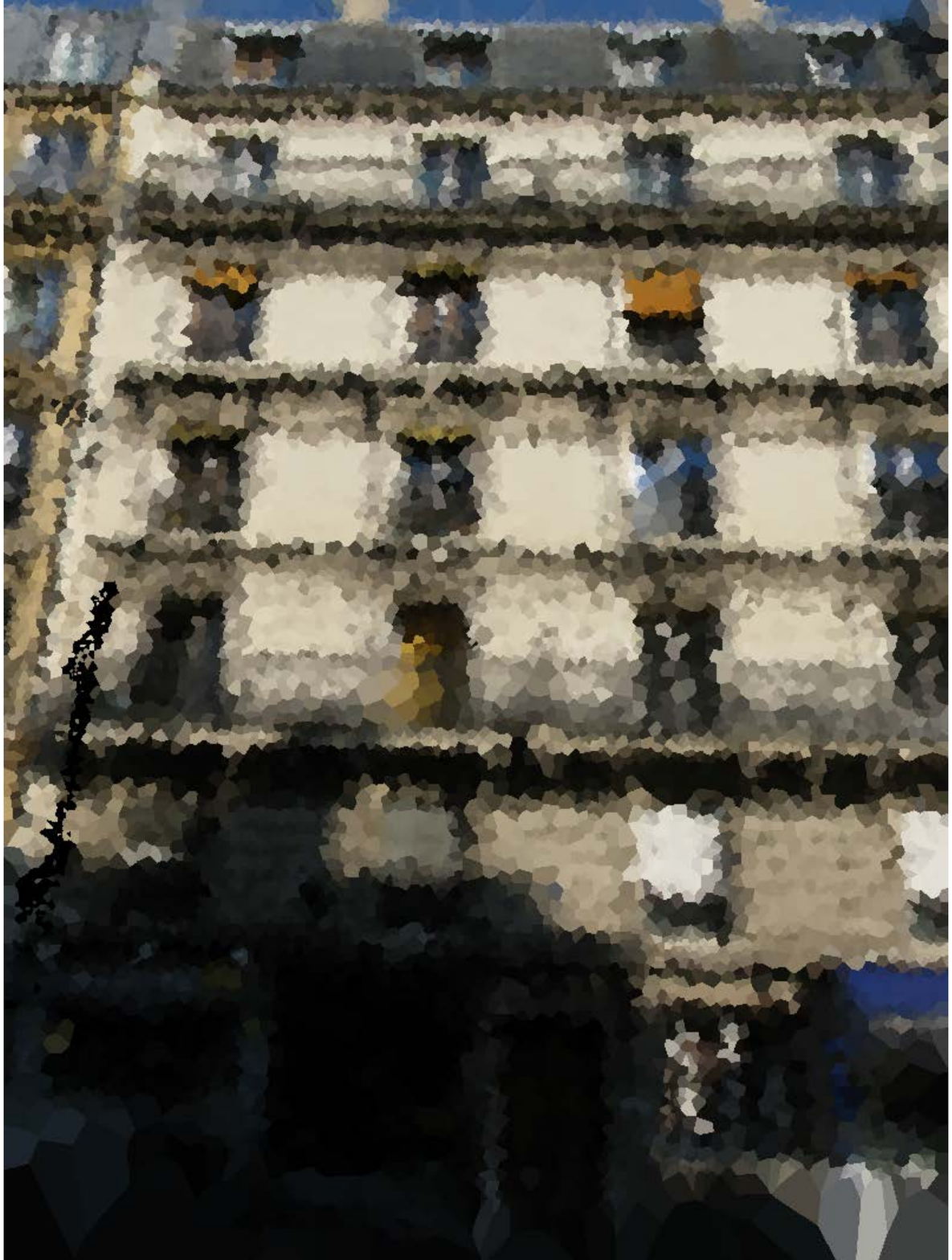} &
    	\includegraphics[height = .14\linewidth, width = .115\linewidth]{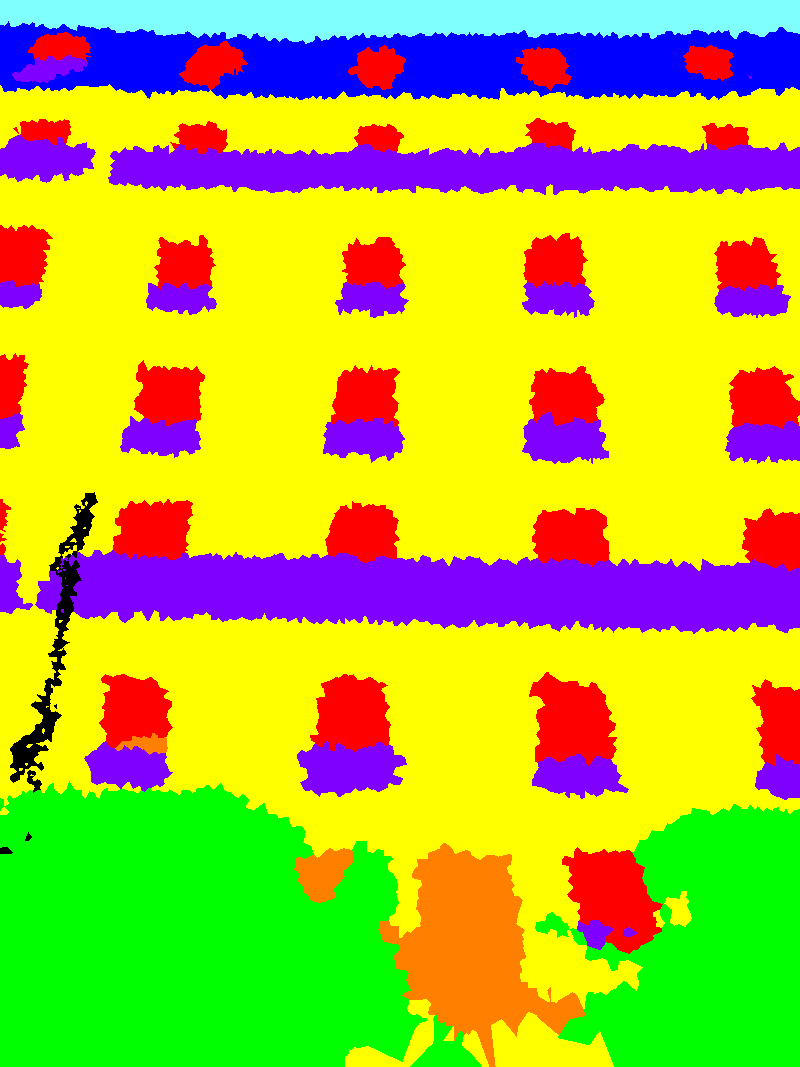} &
    	\includegraphics[height = .14\linewidth, width = .115\linewidth]{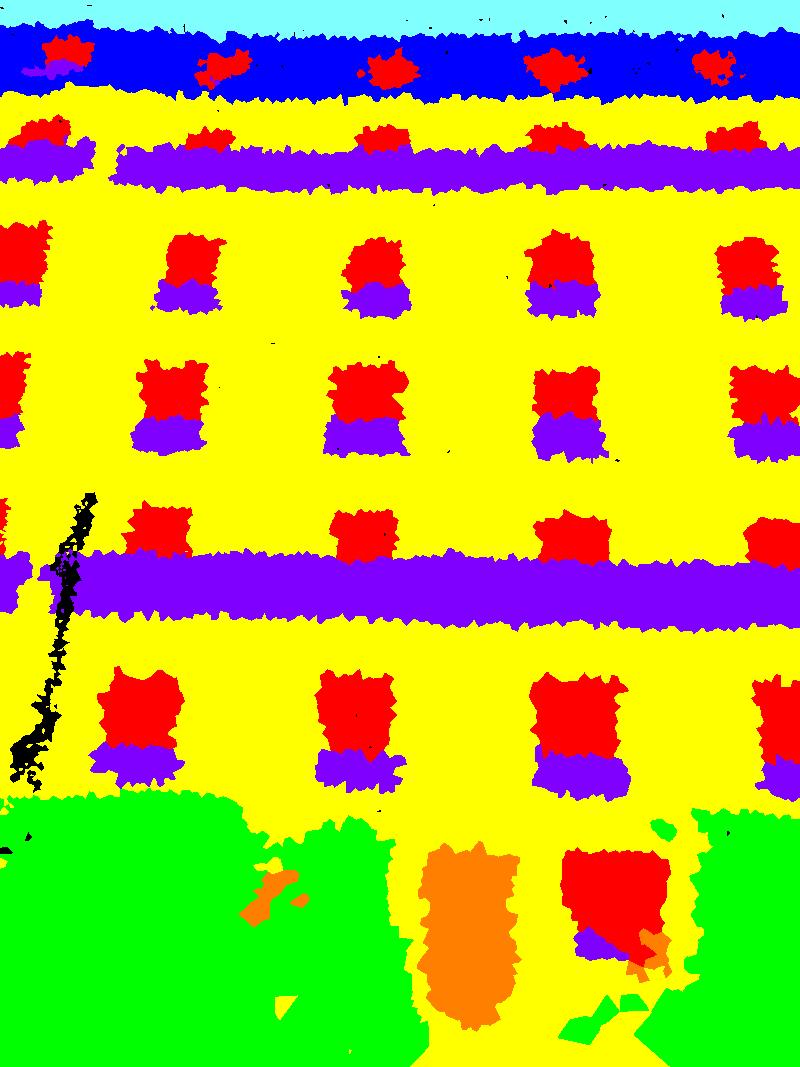} &
    	\includegraphics[height = .14\linewidth, width = .115\linewidth]{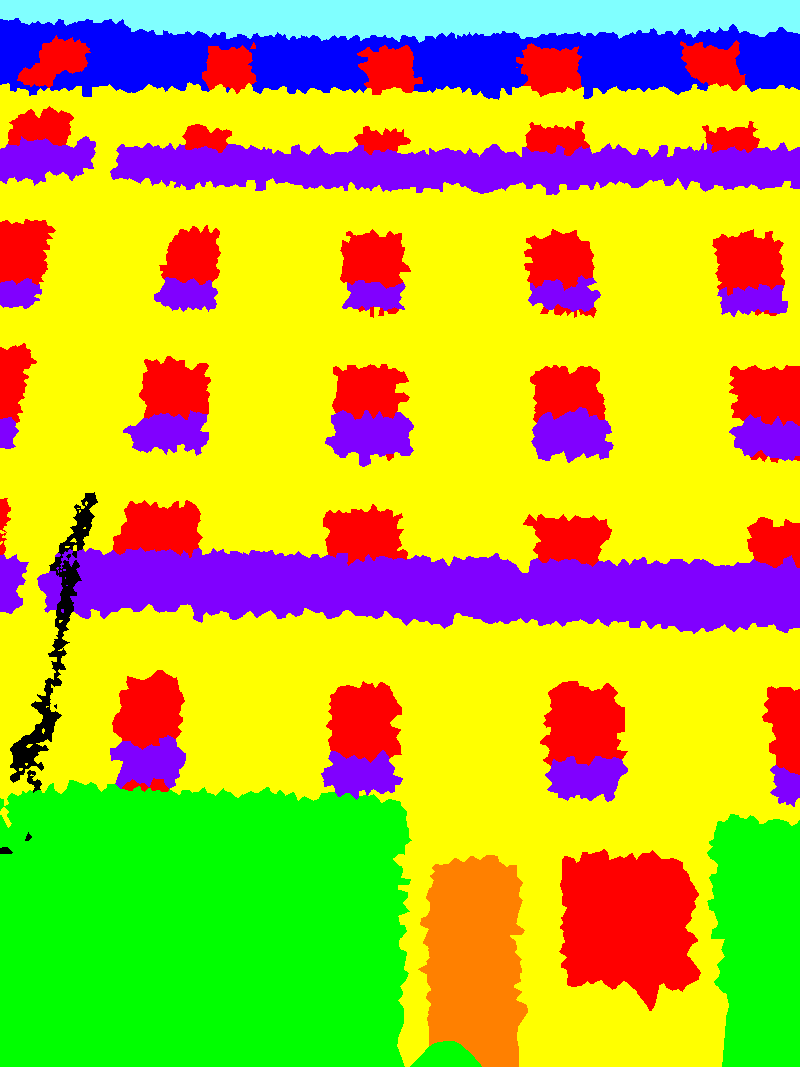}\\
    	(a) input & (b) DRN & (c) +SGPN & (d) GT & (a) input & (b) DRN & (c) +SGPN & (d) GT\\
	\end{tabular}
\caption{A qualitative comparison of different methods for facade segmentation. The results are visualized by mapping the (a) diffuse RGB, (b) predicted and (c) gt label of points to pixels. DRN is the result of directly mapping from the DRN image's segmentation to a point cloud, and SGPN denotes our method (see Section 5.5 in the paper).}
\label{fig:facadeim}
\vspace{-.2cm}
\end{figure*}

\begin{figure*}[h]
    \centering
    \includegraphics[width = 0.99\linewidth]{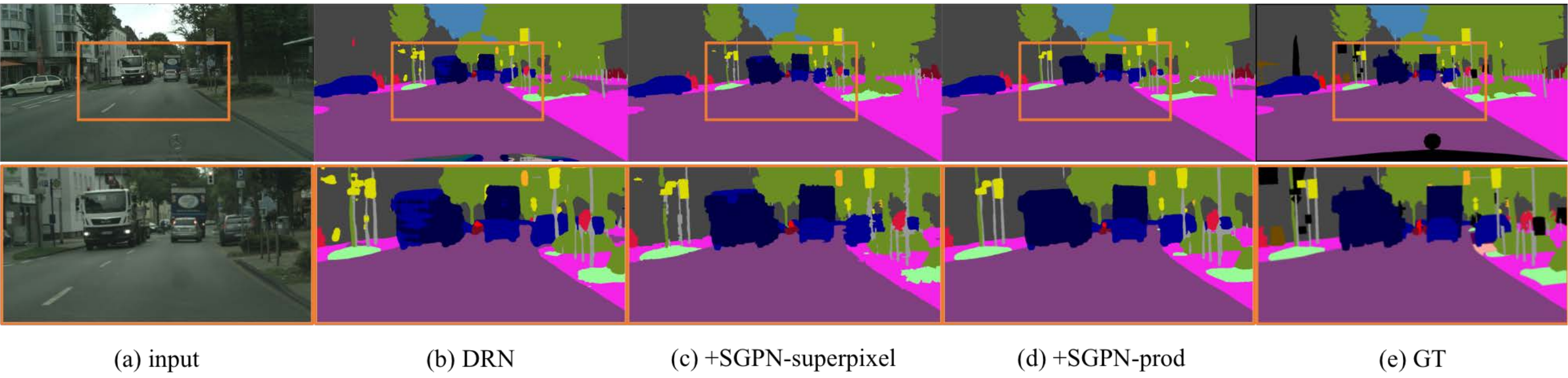}
    \caption{Qualitative comparisons of the Cityscapes semantic segmentation results. (a) DRN-D-22 (baseline); (b) DRN with SGPN on superpixels (see Section 5.4 in the paper); (c) DRN with SGPN on pixels with the inner product kernel (see Section 5.3 in the paper).}
    \label{fig:city}
\end{figure*}

\begin{figure*}[h]
\centering
\begin{tabular}{c@{\hspace{0.01\linewidth}}c@{\hspace{0.01\linewidth}}c@{\hspace{0.01\linewidth}}c@{\hspace{0.01\linewidth}}}
		\includegraphics[width = .24\linewidth]{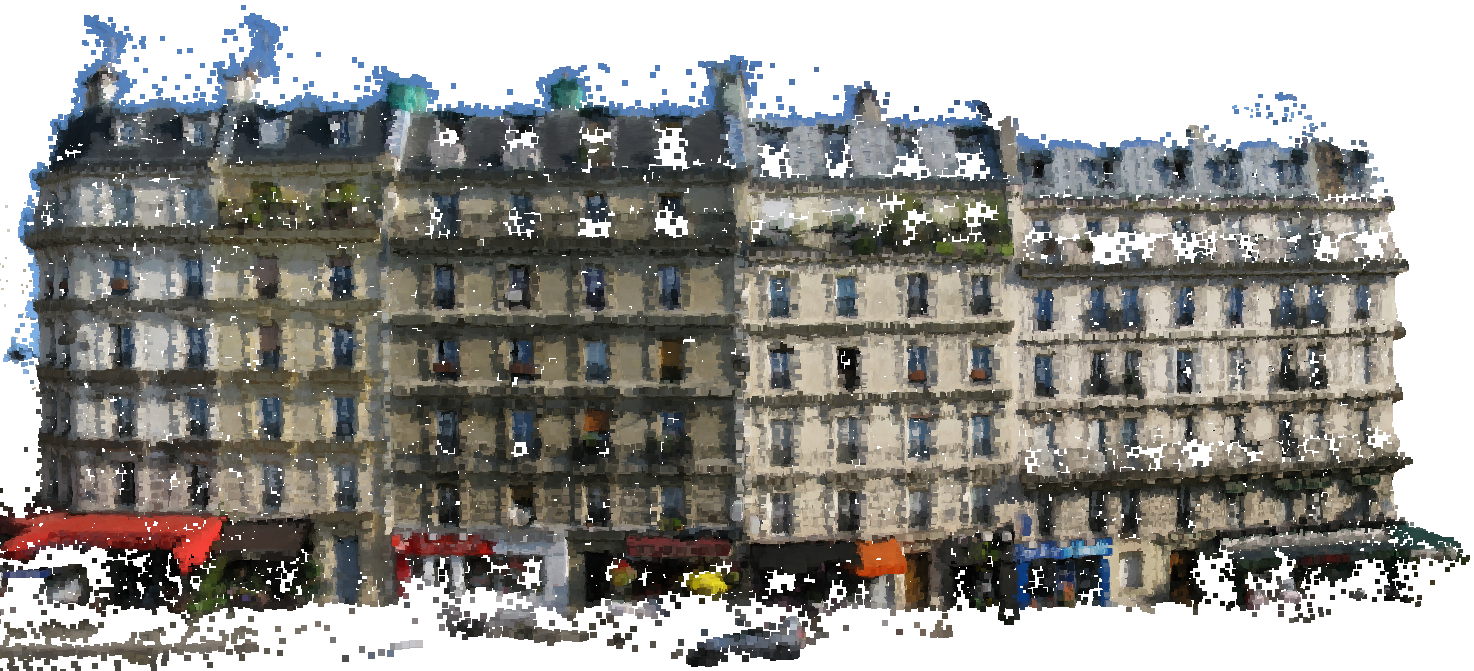} &
    	\includegraphics[width = .24\linewidth]{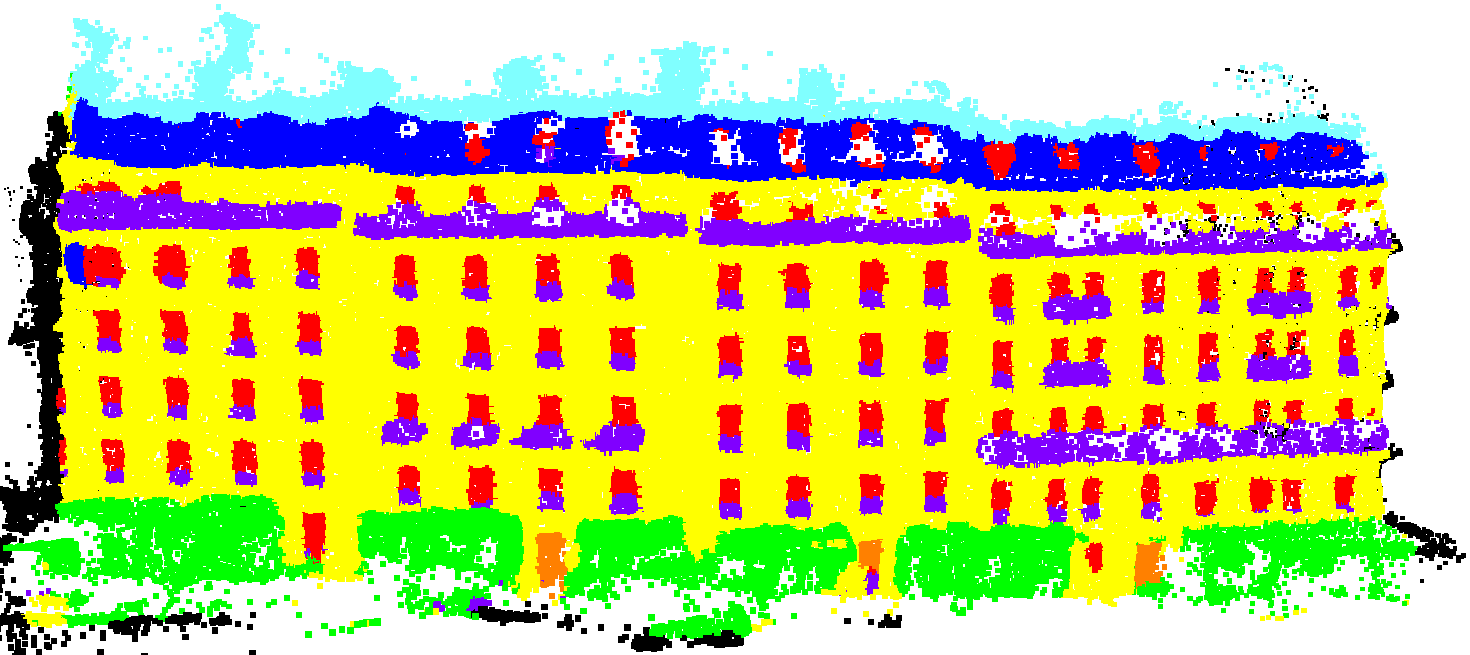} &
    	\includegraphics[width = .24\linewidth]{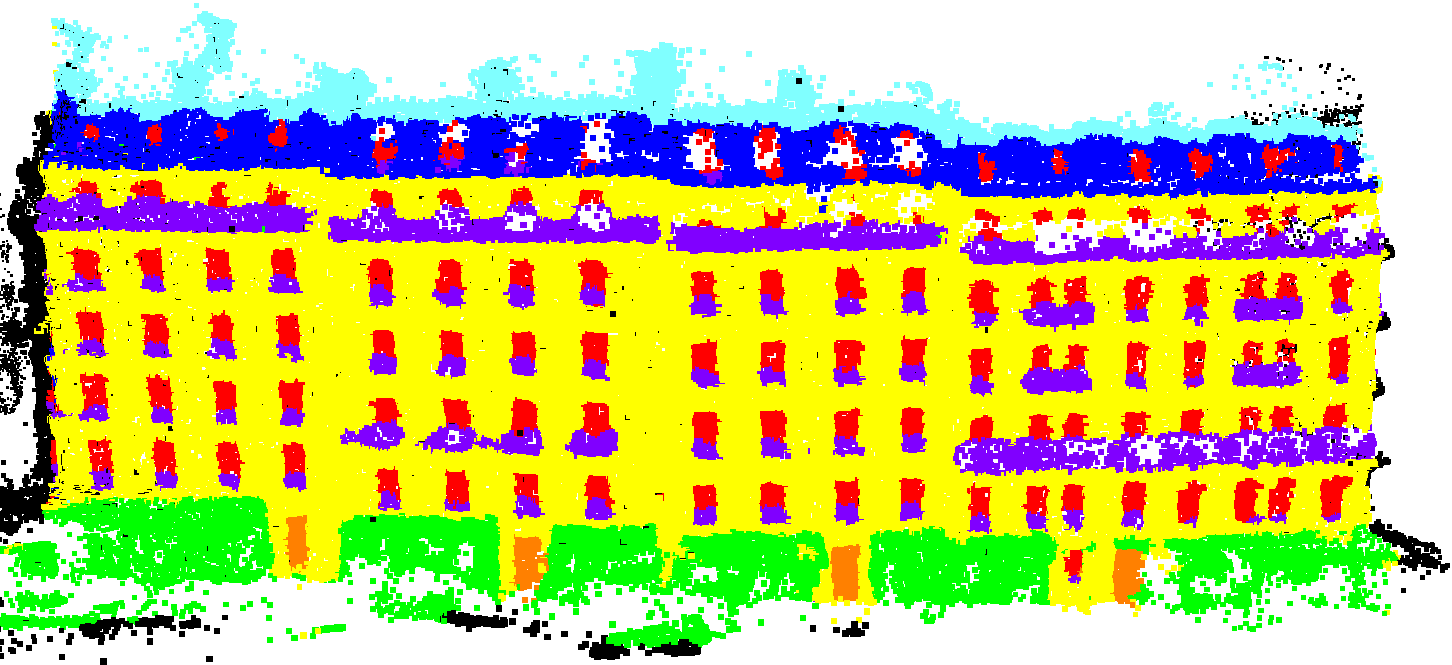} &
    	\includegraphics[width = .24\linewidth]{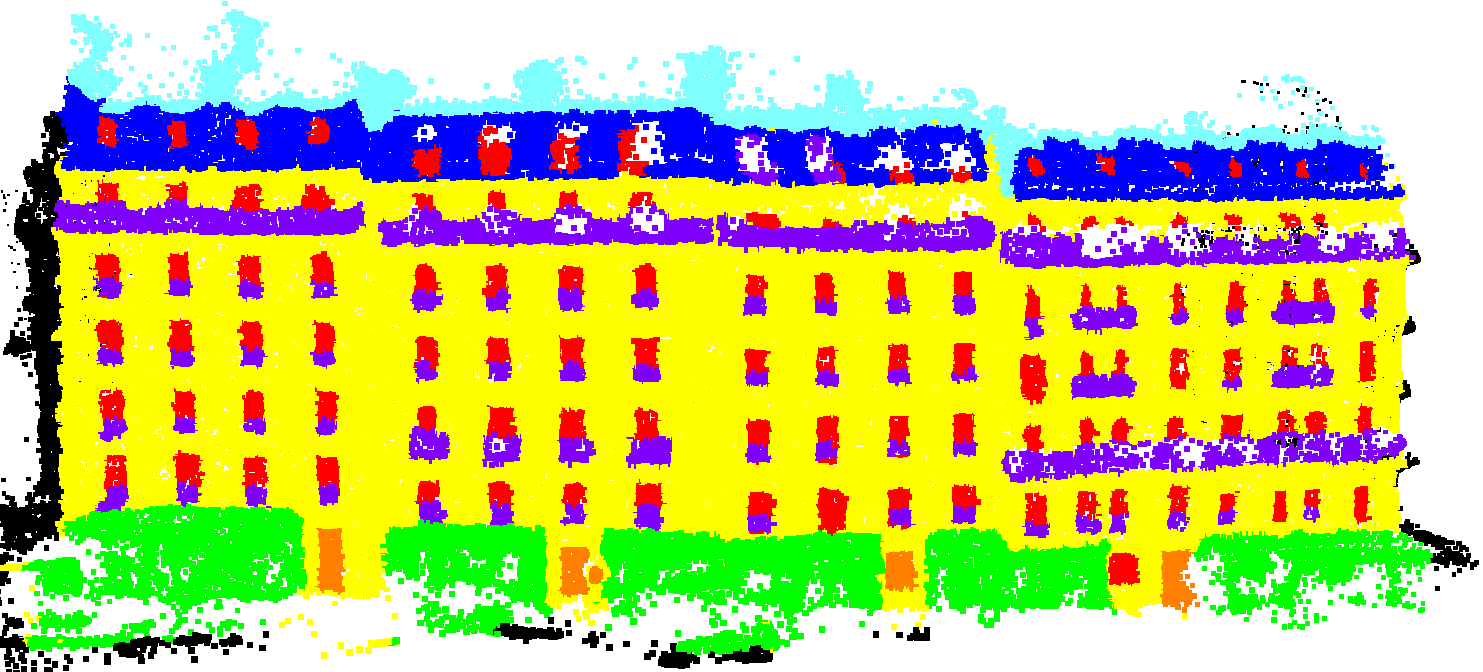} \\
    	(a) image & (b) DRN & (c) +SGPN & (d) GT\\
	\end{tabular}
\caption{Visualized labeling results for (b) DRN as the baseline network and (c) our method (see Section 5.5 in the paper).}
\label{fig:point_cloud}
\vspace{-.2cm}
\end{figure*}

We present more quantitative results of semantic segmentation for the Cityscapes dataset in Table~\ref{tab:drn-pixel}. In addition, we show more qualitative comparisons for the RueMonge~\cite{facade} facade segmentation in Fig.~\ref{fig:facadeim} and \ref{fig:point_cloud}.
We also present the comparison, using the DRN-based network, between the SGPN on superpixels and pixels in Fig.~\ref{fig:city} (c) and (d), respectively.
Note that superpixel-based segmentation shows slightly different appearance due to the local pixel grouping operation. 

\begin{table*}[h]\setlength{\tabcolsep}{3.6pt}
    \centering
    \caption{The full results for \textbf{DRN}- and \textbf{Deeplab}-based networks on the Cityscapes validation set. We show multi-scale testing results except for the +SPN, which shows better performance in single scale setting. \emph{embed} and \emph{prod} denote the embedded Gaussian and inner product kernels. For superpixels, we use the embedded Gaussian kernel since it achieves significantly higher mIoU.}\label{tab:drn-pixel}
    {\footnotesize
    \begin{tabular}{l|lllllllllllllllllll|l}
         categories & \rotatebox[origin=c]{90}{road} & \rotatebox[origin=c]{90}{sidewalk} & \rotatebox[origin=c]{90}{building} & \rotatebox[origin=c]{90}{wall} & \rotatebox[origin=c]{90}{fence} & \rotatebox[origin=c]{90}{pole} & \rotatebox[origin=c]{90}{trafficlight} & \rotatebox[origin=c]{90}{trafficsign} & \rotatebox[origin=c]{90}{vegetation} & \rotatebox[origin=c]{90}{terrain} & \rotatebox[origin=c]{90}{sky} & \rotatebox[origin=c]{90}{person} & \rotatebox[origin=c]{90}{rider} & \rotatebox[origin=c]{90}{car} & \rotatebox[origin=c]{90}{truck} & \rotatebox[origin=c]{90}{bus} & \rotatebox[origin=c]{90}{train} & \rotatebox[origin=c]{90}{motorcycle} & \rotatebox[origin=c]{90}{bicycle} & mIoU \\\hline
        Baseline DRN~\cite{Yu2017} & 97.4 & 80.9 & 91.1 & 32.9 & 54.9 & 60.6 & 65.6 & 75.9 & 92.1 & 59.1 & 93.4 & 79.2 & 57.8 & 92.0 & 42.9 & 65.2 & 55.2 & 49.4 & 75.2 & 69.5 \\
        +SPN (single)~\cite{liu2017learning} & 97.7 & 82.7 & 91.3 & 34.4 & 54.6 & 61.8 & 65.9 & 76.4 & 92.2 & 62.2 & 94.4 & 78.5 & 56.8 & 92.7 & 47.8 & 70.2 & 63.6 & 52.0 & 75.3 & 71.1 \\
        +NLNN~\cite{Wang_nonlocalCVPR2018} & 97.4 & 81.1 & 91.2 & 43.0 & 52.9 & 60.3 & 66.1 & 74.9 & 91.7 & 60.6 & 93.4 & 79.2 & 57.7 & 93.3 & 54.4 & 73.5 & 54.7 & 54.2 & 74.4 & 71.3 \\\hline
        +SGPN-embed & 98.0 & 83.8 & 92.2 & 48.5 & \textbf{59.7} & 64.1 & 70.1 & 79.4 & 92.6 & \textbf{63.8} & \textbf{94.7} & 82.0 & 60.7 & 94.9 & 62.5 & 77.7 & 51.1 & \textbf{62.8} & \textbf{77.6} & 74.5 \\
        +SGPN-prod & \textbf{98.1} & \textbf{84.4} & \textbf{92.2} & 51.8 & 56.5 & \textbf{65.8} & \textbf{71.2} & \textbf{79.4} & \textbf{92.7} & 63.2 & 94.3 & \textbf{82.7} & \textbf{65.1} & \textbf{94.9} & \textbf{73.8} & 78.0 & 43.2 & 59.7 & 77.4 & \textbf{75.0} \\
        +SGPN-superpixels & 97.6  & 82.4 & 91.0 & \textbf{52.7} & 52.9 & 58.4 & 66.1 & 75.9 & 91.8 & 62.2 & 93.6 & 79.4 & 58.2 & 93.3 & 62.4 & \textbf{79.7} & 57.1 & 60.2 & 75.1 & 73.2\\\hline\hline
        Baseline Deeplab~\cite{deeplabv3plus2018} & 98.0 & 85.0 & 93.0 & 52.9 & 63.6 & 67.0 & 72.6 & 80.0 & 92.4 & 56.8 & 95.0 & 84.1 & 67.3 & 95.4 & 84.0 & 88.5 & 78.5 & 67.8 & 78.5 & 79.0 \\
        +SGPN-embed & \textbf{98.3} & \textbf{85.4} & 93.1 & \textbf{54.6} & 62.0 & 69.3 & 74.0 & \textbf{82.6} & \textbf{92.7} & 63.4 & 95.0 & 85.0 & 69.2 & 95.9 & \textbf{85.5} & 89.3 & 85.3 & 69.5 & 78.0 & 80.4 \\
        +SGPN-prod & 98.1 & 85.2 & \textbf{93.3} & 50.6 & \textbf{65.8} & \textbf{70.7} & \textbf{74.9} & 81.5 & 92.6 & 63.4 & \textbf{95.2} & \textbf{85.1} & \textbf{69.3} & \textbf{96.0} & 85.0 & \textbf{91.8} & \textbf{86.7} & \textbf{71.8} & \textbf{80.1} & \textbf{80.9} \\\hline
    \end{tabular}
    }
\end{table*}

\subsection{Color Restoration and Scribble Propagation}

\paragraph{Color restoration.}
We show that the SGPN itself is general and can be used for diffusing different type of properties, \eg color, HDR and masks.
Here we conduct \emph{sparse color restoration} using SGPN:
In the Lab color space, we randomly mask out pixels in the \textit{ab} channels. The SGPN is used to propagate the color information from the sparse colored pixels to the non-colored regions purely based on the affinities learned from the L channel.
We use a light-weight CNN and the dot-product kernel to compute strengths of the edges of our graph (the orange links in Fig.~4 in paper), while the unary feature (the green nodes in Fig.~4 in paper) is replaced by the sparse color pixels.
We train a single network where we retain only $1\%\sim3\%$ of randomly selected colored pixels during training. During testing, we show that the model can generalize well to different proportions of randomly selected colored pixels, e.g., $[1\%, 5\%, 10\%, 20\%]$, as shown in Fig.~\ref{fig:01}.
The results reveal that the SGPN: (a) successfully learns general affinity and does not overfit to a certain ratio of the available colored pixels, and (b) the propagation is global so that very sparse colored pixels (e.g., $1\%$ in Fig.~\ref{fig:01} (a)) can ``travel long distances'' to fill-in the whole image with color.

\begin{figure*}[h]
    \centering
    \footnotesize
\begin{tabular}{c@{\hspace{0.005\linewidth}}c@{\hspace{0.005\linewidth}}c@{\hspace{0.005\linewidth}}c@{\hspace{0.005\linewidth}}c@{\hspace{0.005\linewidth}}}
    \includegraphics[width=.19\linewidth]{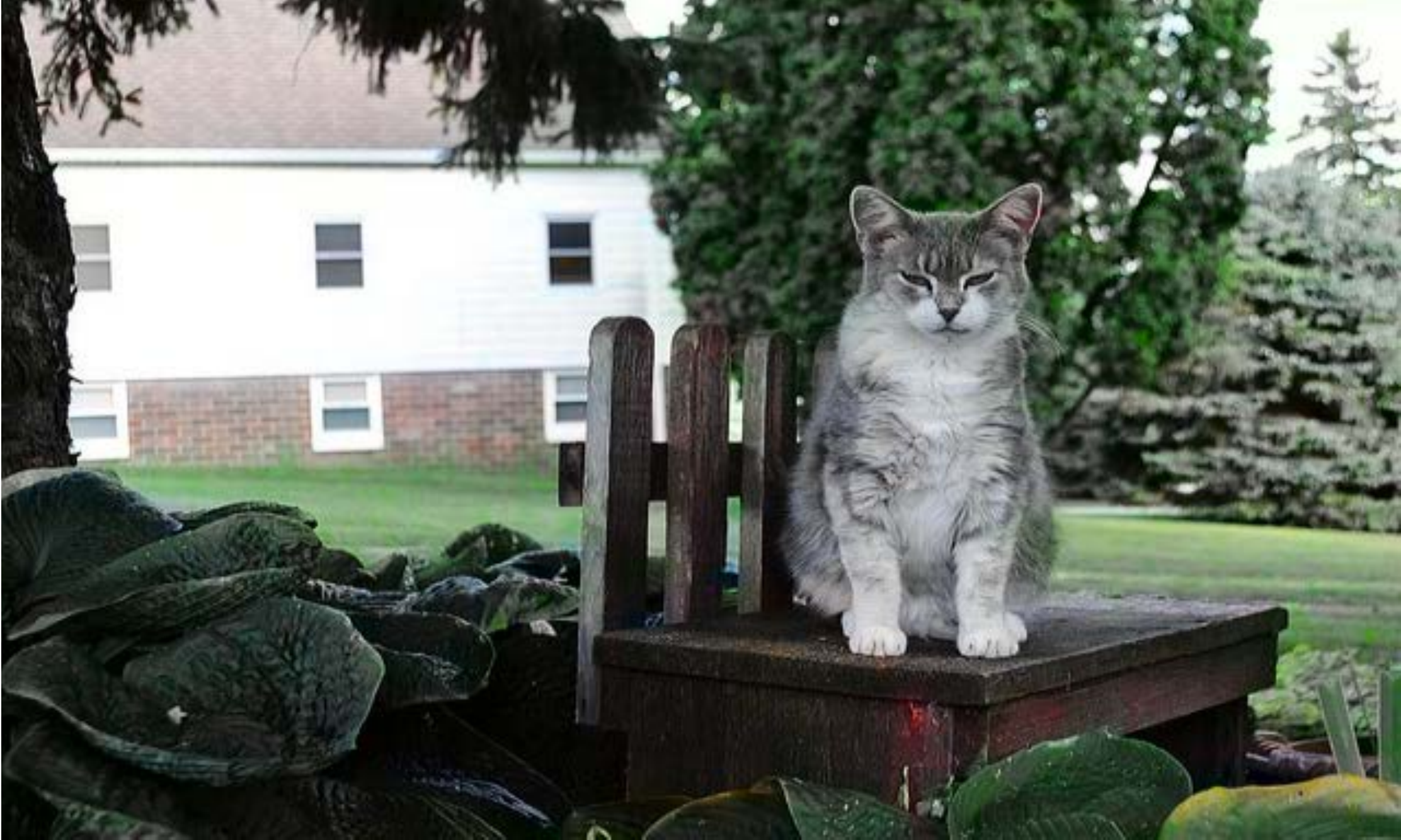} &
    \includegraphics[width=.19\linewidth]{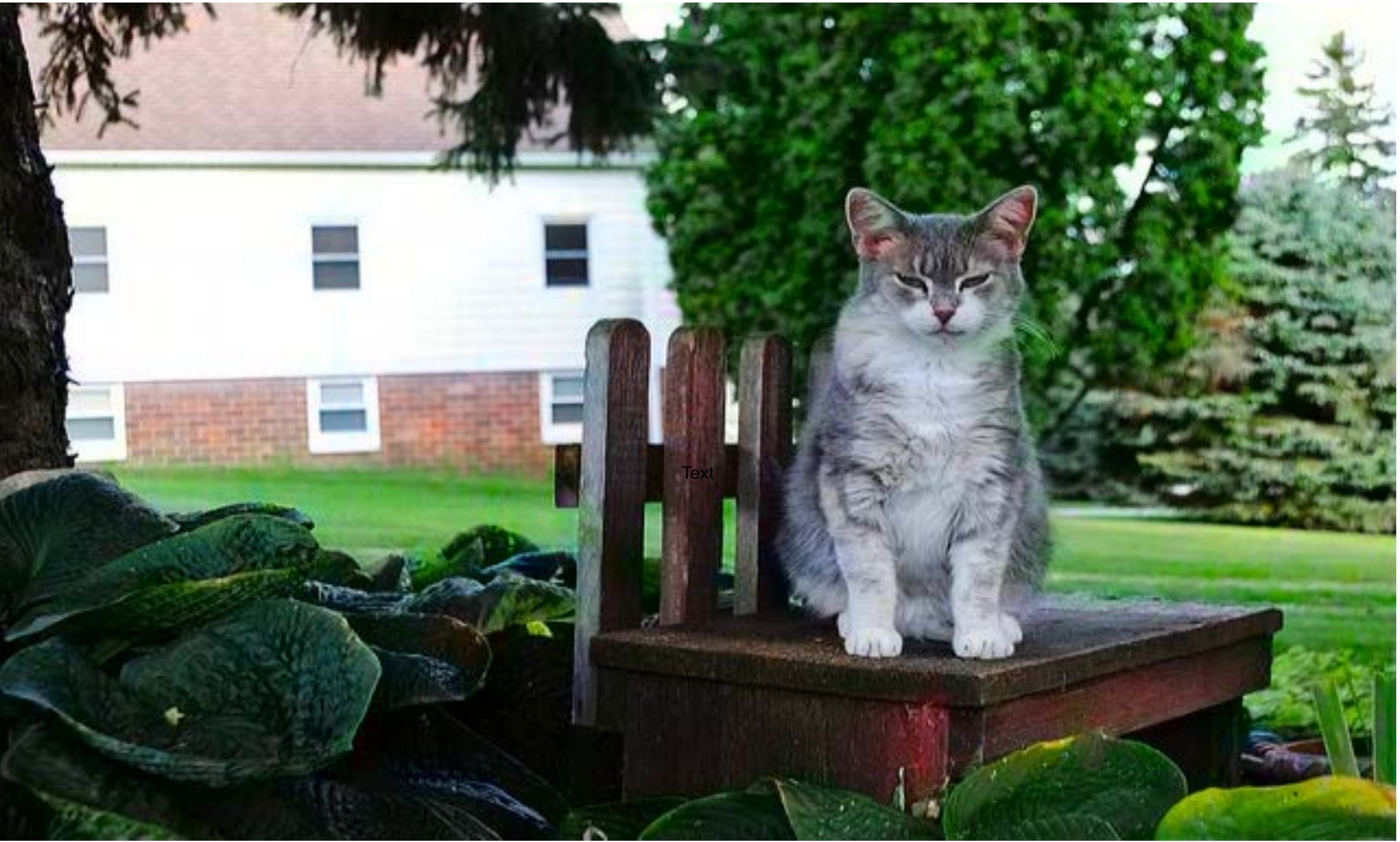} &
    \includegraphics[width=.19\linewidth]{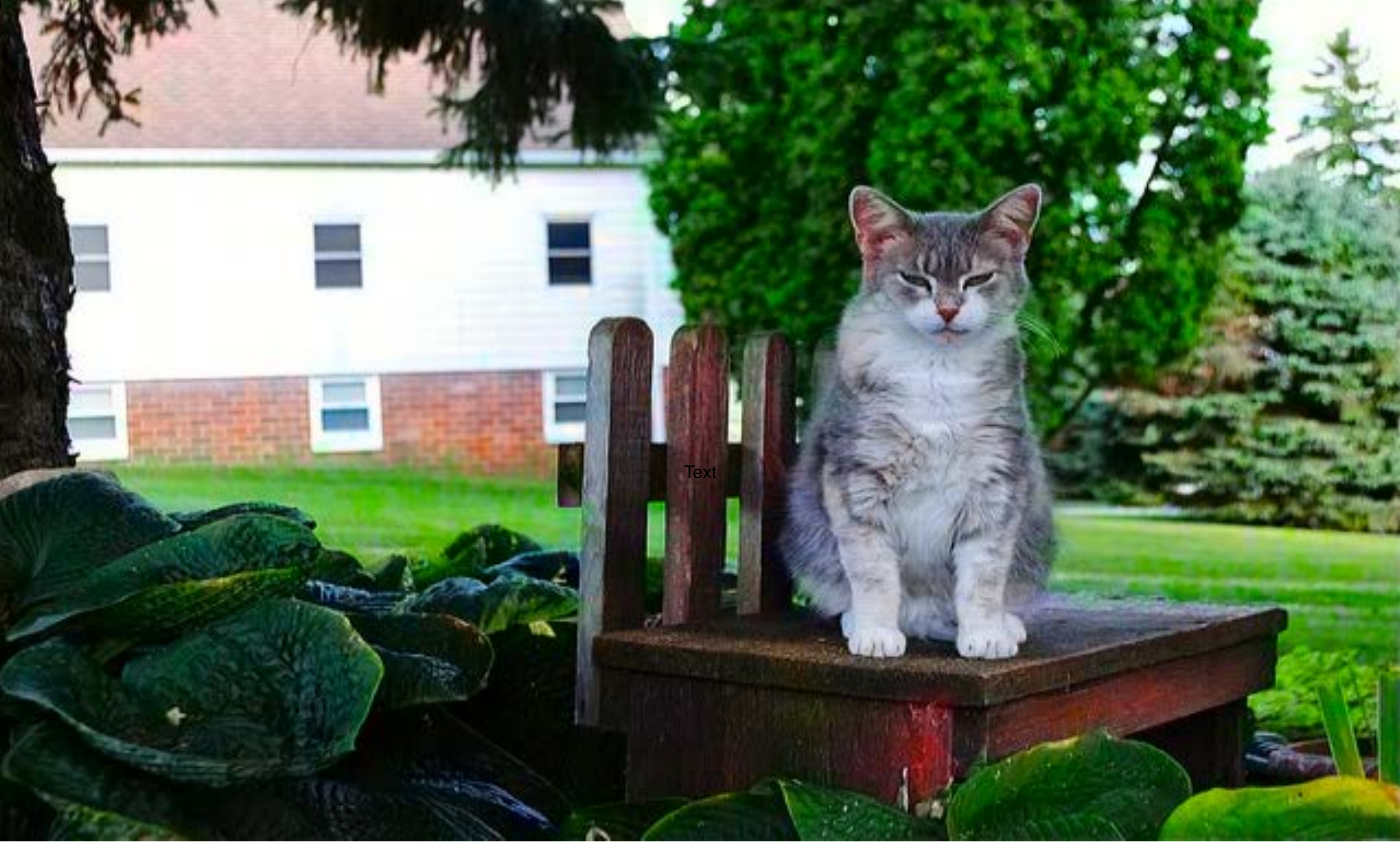} &
    \includegraphics[width=.19\linewidth]{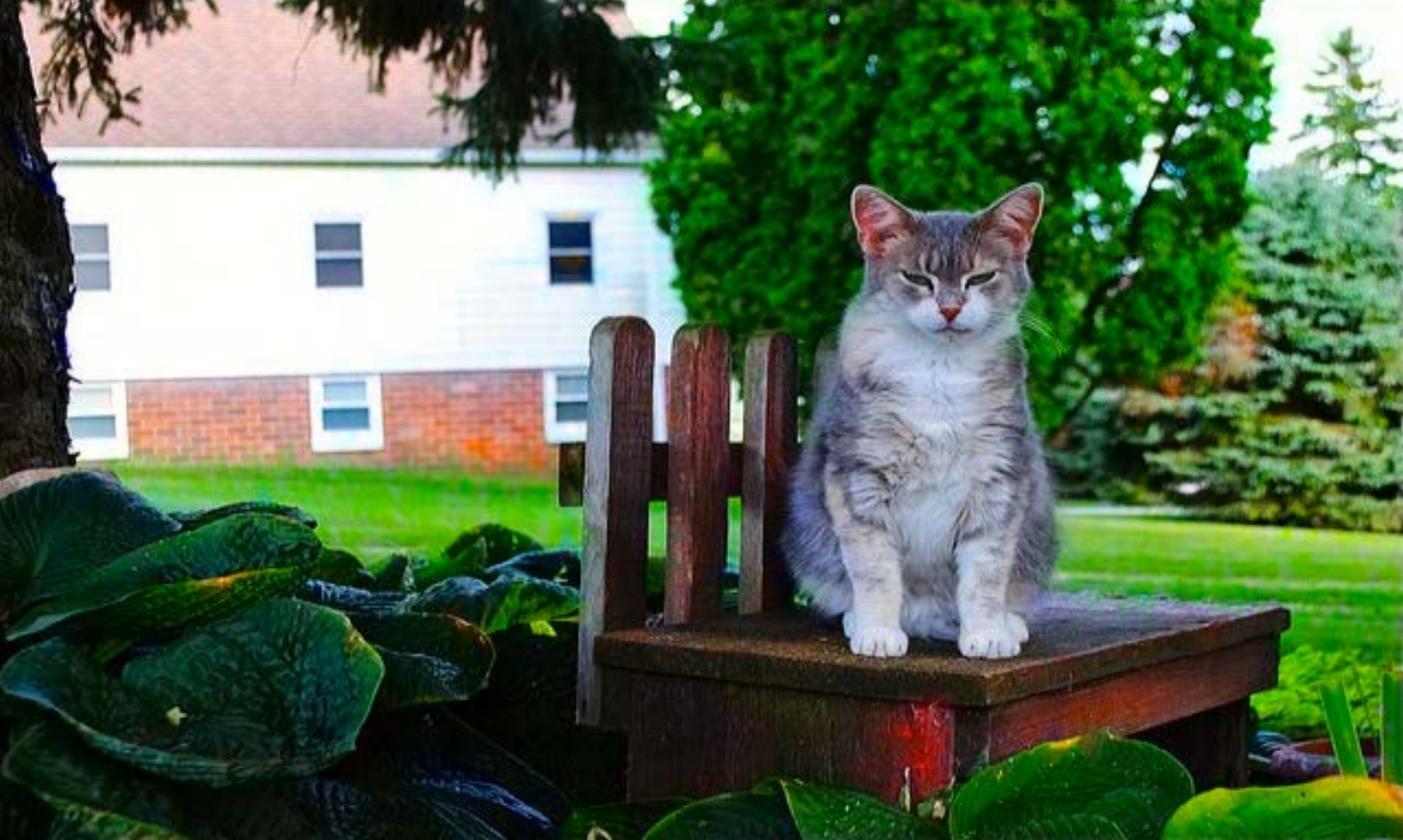} &
    \includegraphics[width=.19\linewidth]{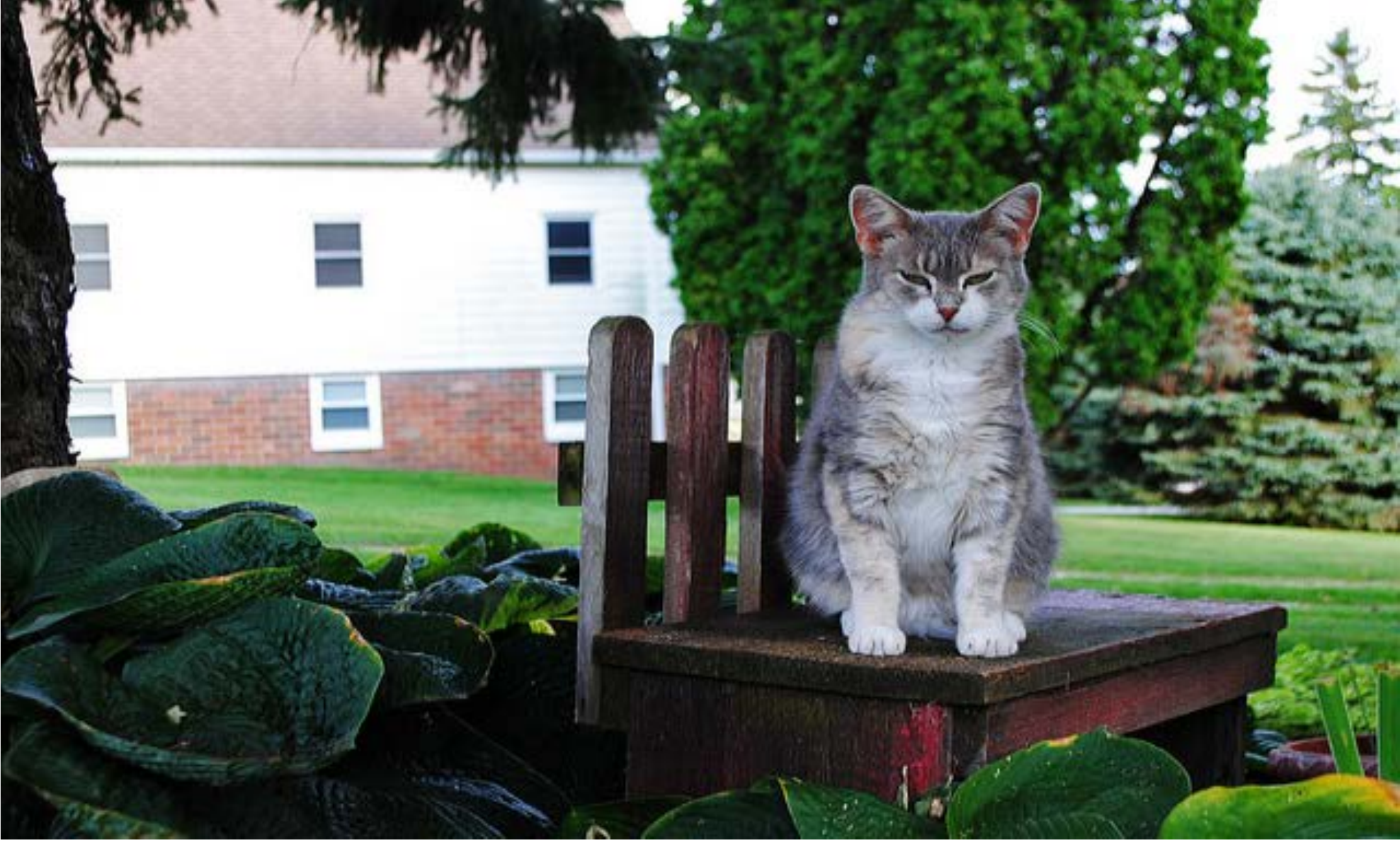} \\
    (a) $1\%$ & (b) $5\%$ & (c) $10\%$ & (d) $20\%$ & (e) ground truth\\
\end{tabular}
    \caption{Color restorations with different ratios (a-d) of colored pixels retained from the original image (e).}
    \label{fig:01}
\end{figure*}

\begin{figure*}[h]
    \centering
    \includegraphics[width = 0.99\linewidth]{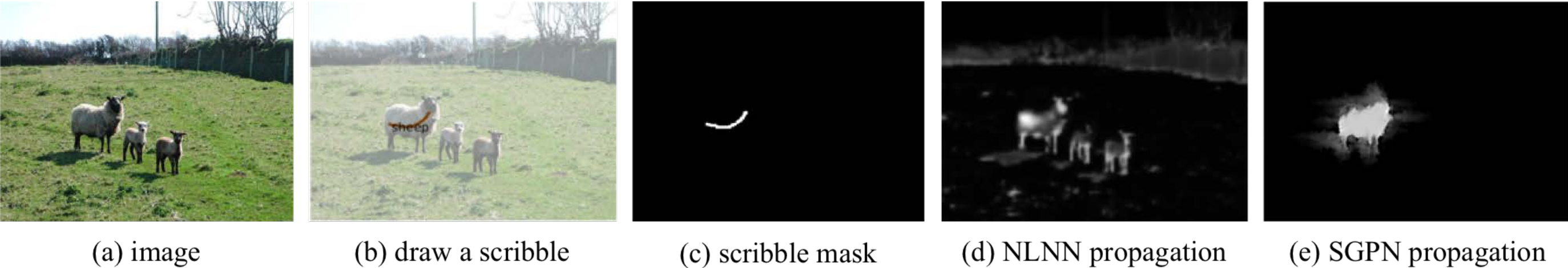}
    \caption{Scribble propagation via affinities learned for the task of color restoration. Different propagation modules are utilized, including (d) NLNN~\cite{Wang_nonlocalCVPR2018} and (e) SGPN as proposed in the paper.}
    \label{fig:scribble}
\end{figure*}

\paragraph{Scribble propagation.}
In addition, we show that the learned affinity from the task of color restoration, can be directly used for propagation of other properties as well, such as scribbles.
We show an example of this in Fig.~\ref{fig:scribble}, where a single scribble is drawn on one of the sheep in the image.
With the same model trained that we trained for color restoration, but with a scribble mask as the ``unary" feature to propagate (see Fig.~\ref{fig:scribble} (c)) instead of the color channels (we duplicate the mask to fit to the input dimension of the network), our SGPN can faithfully propagate to spatial regions ranging in shape from a thin line, to the whole body of the sheep, while maintaining the fine structures and details, such as the legs of the sheep (see Fig.~\ref{fig:scribble} (e)).

Additionally, we also train a color restoration network with NLNN~\cite{Wang_nonlocalCVPR2018}, with a similar strategy: we append a non-local module to a CNN, which takes gray-scale images as input to learn the non-local affinity for color propagation. Different from the proposed spatial propagation module, in NLNN, the restored color map is obtained by directly multiplying the affinity matrix with the sparse color maps.
We adopt the first 4 layers of ResNet18~\cite{he2016deep} with pretrianed weights from ImageNet  as the CNN (the light-weight CNN trained from scratch for our SGPN does not converge for NLNN) before the non-local module. Similar to SGPN, we directly apply the learned affinity to propagate the same scribble mask (see Fig.~\ref{fig:scribble} (c)).
Due to the non-local nature of the algorithm, the scribble is propagated globally to all the other regions with similar colors as the annotated sheep, which results in a much more noisy mask (see Fig.~\ref{fig:scribble} (d)), compared with SGPN.
This is also a real-world example revealing the importance of path-aware propagation, as we explained in Fig.~1 in the paper.